\documentclass[preprint,8pt]{elsarticle}
\usepackage{hyperref}
\usepackage{graphicx}
\usepackage{float}
\usepackage{subfig}
\usepackage{makecell}
\usepackage{amsmath}
\usepackage{amssymb}
\usepackage{booktabs}
\usepackage{multirow}
\usepackage{color}
\usepackage{algorithm} 
\usepackage{algorithmic} 

\journal{Journal of Optik}

\usepackage{amssymb}

\begin{document}

\begin{frontmatter}
	\title{Metropolis Theorem and Its Applications in Single Image Detail Enhancement}
	
	\author[ind1]{He Jiang\corref{cor1}}
	\author[ind2]{Mujtaba Asad}
	\author[ind1]{Jingjing, Liu}
	\author[ind1]{Haoxiang, Zhang}
	\author[ind1]{Deqiang Cheng\corref{cor2}}
	
	\cortext[cor1]{Corresponding author 1: He Jiang,  E-mail address: \textcolor{cyan}{jianghe@cumt.edu.cn}}
	\cortext[cor2]{Corresponding author 2: Deqiang Cheng,  E-mail address: \textcolor{cyan}{chengdq@cumt.edu.cn}}
	
	\address[ind1]{School of Information and Control Engineering, China University of Mining and Technology, Xuzhou, China.}
	\address[ind2]{Department of Computer Science, University of Central Punjab, Lahore Pakistan }
	
	\begin{abstract}
		Traditional image detail enhancement is local filter-based or global filter-based. In both approaches, the original image is first divided into the base layer and the detail layer, and then the enhanced image is obtained by amplifying the detail layer. Our method is different, and its innovation lies in the special way to get the image detail layer. The detail layer in our method is obtained by updating the residual features, and the updating mechanism is usually based on searching
		and matching similar patches. However, due to the diversity of image texture features, perfect matching is often not possible. In this paper, the process of searching and matching is treated as a thermodynamic process, where the Metropolis theorem can minimize the internal energy and get the global optimal solution of this task, that is, to find a more suitable feature for a better detail enhancement performance. Extensive experiments have proven that our algorithm can achieve better results in quantitative metrics testing and visual effects evaluation. The source code can be obtained from the
		\href{https://github.com/hehesjtu/Metropolis-Theorem}{link}.
	\end{abstract}
	
	
\end{frontmatter}

\section{Introduction}
Image detail enhancement algorithm has been widely used. With the popularity of consumer electronics, thousands of images are created. Due to the limitations of the equipment and the complexity of the environment, they are often accompanied by blurring and noise, thus a robust and effective detail enhancement algorithm is desperately needed.

The first thing to note here is that the image enhancement is not the same as the image detail enhancement. Firstly, in terms of applications, image enhancement algorithms are mostly used for low-quality image processing, while image detail enhancement algorithms can be used for images of any quality. Secondly, image enhancement increases information entropy by superimposing information that does not belong to the original signal itself. Image detail enhancement is to decompose the image into a smooth layer and a detail layer. By enlarging the detail layer, the resultant signal has a better visual performance. Therefore, image detail enhancement does not increase the information entropy of the original signal.

Image detail enhancement and image smoothing are essentially the same. Both algorithms divide the original image $ I $ into a combination of a smooth layer $ \mathcal{S} $ and a detail layer $ \mathcal{D} $, where $ I = \mathcal{S} + \mathcal{D} $ and $ \mathcal{S} = f(I) $. 
$ f(*) $ is a pixel-wise filter that extracts the smooth layer from the original image $ I $. Since $ I $ is a priori signal, the image smoothing and detail enhancement algorithms are equivalent. According to the different design ways of the filter $ f(*) $, algorithms can be classified into local filter-based methods and global filter-based methods. 

Local filters, such as median filter, bilateral filter [\textcolor{cyan}{20}], and Guided Image Filter (GIF) [\textcolor{cyan}{1}, \textcolor{cyan}{2}] are pioneering work. They run fast, but defects
exist in their enhanced images, namely jagged defect, gradient reversal defect, and halo defect, respectively. Many subsequent filters [\textcolor{cyan}{3}, \textcolor{cyan}{4}, \textcolor{cyan}{7}, \textcolor{cyan}{8}, \textcolor{cyan}{14}, \textcolor{cyan}{22}, \textcolor{cyan}{23}, \textcolor{cyan}{24}, \textcolor{cyan}{25}, \textcolor{cyan}{30}, \textcolor{cyan}{31}] improve some aspects of these three filters to get better performance. RGF (Rolling Guide Filter) [\textcolor{cyan}{7}] greatly solves the problem of image edge being incorrectly smoothed by finding the way of scale perception. GGIF [\textcolor{cyan}{3}] (Gradient-domain Guided Image Filter)
weights the GIF via image gradient features to get a clearer result. EGIF [\textcolor{cyan}{8}] (Effective Guided Image Filter)
adopts a mechanism to get an adaptive detail magnification factor. SPGIF [\textcolor{cyan}{14}] (Structure-Preserved Guided Image Filter) introduces a global constraint based on GIF to suppress noise while preserving image structure. However, the results of RGF [\textcolor{cyan}{7}], GGIF [\textcolor{cyan}{3}], and EGIF [\textcolor{cyan}{8}] and  sometimes are excessively enhanced. Besides, the result of the SPGIF [\textcolor{cyan}{14}] can sometimes change the chroma of the image, making it visually a bad experience. 

In addition to such local filters, there are global filters that can do the job. A Weighted Least Square (WLS) filter [\textcolor{cyan}{11}] is proposed, which considers the global information and builds the optimization equation to solve the problem.  Many subsequent methods [\textcolor{cyan}{10}, \textcolor{cyan}{12}, \textcolor{cyan}{13}, \textcolor{cyan}{19}, \textcolor{cyan}{21}, \textcolor{cyan}{26}, \textcolor{cyan}{33}] have been proposed on the basis of WLS [\textcolor{cyan}{11}]. These methods differ in a way that they use superior measures to model the global information or construct different optimization techniques to solve this problem. For example, Fractal Set (FS) [\textcolor{cyan}{13}] is used as the measure of the image detail layer, and the authors enhance the image detail layer by maximizing the fractal length.  BFLS [\textcolor{cyan}{21}] embeds the Bilateral Filter into the Least Square model for effective image detail boosting.  Furthermore, some more mathematically complex penalty terms are designed, i.e. Iterative Least Square (ILS) [\textcolor{cyan}{10}] or Truncated Huber (TH) [\textcolor{cyan}{33}], which are applied as the penalizing terms to optimize the equation. Because more features are referenced, the enhancement performance is better. However, the running speed of these approaches is slow.

In general, the algorithms based on the local filter and global filter can be written in Eq. 1, where $ \mathcal{S} $ and $ I $ are the output image and the input
image, respectively, $ \mathcal{F}(\mathcal{S}, I) $ is the fidelity term between $ \mathcal{S} $ and $ I $, $ \mathcal{R}(\mathcal{S}) $ is a regular term designed according to specific requirements, and $ \lambda $ is a non-negative constant used to balance $ \mathcal{F}(\mathcal{S}, I) $ and $ \mathcal{R}(\mathcal{S}) $. The final detail enhancement image $ I_{enhanced} $ can be computed by Eq. 2, and $ \alpha $ is the detail layer magnification factor. 
\begin{equation}
	\mathcal{S} = \arg\min \mathcal{F}(\mathcal{S},\mathcal{I}) + \lambda \mathcal{R}(\mathcal{S})
\end{equation}
\begin{equation}
	I_{enhanced} = I + \alpha \times (I-\mathcal{S})
\end{equation}

Apart from these two traditional methods,  another main method, for image detail enhancement, that has emerged recently is based on residual learning.
The detail enhancement algorithm based on residual learning is first proposed in [\textcolor{cyan}{27}]. Unfortunately, this algorithm is only applicable to wide-angle images, which undoubtedly limits its generalization. In [\textcolor{cyan}{9}], the author adopts Zero-order Filter (ZF) to fit residual features, but the low order makes its fitting ability limited. In view of this, the author of [\textcolor{cyan}{5}, \textcolor{cyan}{6}] proposes a detail enhancement algorithm based on In-Place Residual Homogeneity (IPRH) and obtains the image detail layer by means of searching and matching. However, the searching process is a greedy mechanism, which makes the algorithm converge to the local optimal solution with a high probability.

In our proposed work, we still use searching and matching technique to get the residual layer, that is, the rough detail layer, but unlike the methods in [\textcolor{cyan}{5}, \textcolor{cyan}{6}], the searching and matching process will be analogous to the process of cooling a thermodynamic system. With the help of the Metropolis theorem, a thermodynamic system can find where the lowest point of internal energy locates. This way of obtaining low values of energy will be learned in our system so that the system can converge to the global optimal solution. 

The detail enhancement algorithm based on residual learning can be described by Eq. 3 $ \sim $ Eq. 5. As shown in Eq. 3 and Eq. 4, $ \mathbf{s}(\mathbf{x}) $ is the coordinates of offsets, $ \mathbf{x} $ is the position of a patch, and $ \mathcal{P}(\mathbf{x}) $ is a patch centered at $ \mathbf{x} $ in image $ I $, and $ \Omega $ is the feasible region of $ \mathbf{x} $. By computing all the image patches, the initial residual feature $ Res $ is obtained. Due to the roughness of the initial residual feature, it is necessary to design a mechanism $ f(*) $ to update this feature. ZF[\textcolor{cyan}{9}] updates the residual feature $ Res $ by zero-order filtering. RH [\textcolor{cyan}{5}, \textcolor{cyan}{6}] refines $ Res $ through searching and matching, and finally obtains the detail layer $ f(Res) $, where $ I_{enhanced} $ in Eq. 5 is the final detail enhancement image, and $ \alpha $ is the detail layer magnification factor.  In terms of subjective performance, RH [\textcolor{cyan}{5}, \textcolor{cyan}{6}] is better than ZF[\textcolor{cyan}{9}]. However, it is almost impossible to obtain the
global optimum by only using RH’s [\textcolor{cyan}{5}, \textcolor{cyan}{6}] method to update the residual feature. Therefore, it is necessary to design a new residual update mechanism $ f_{new}(*) $,  and the novelty of this paper comes from this.

\begin{equation}
	\mathbf{s}(\mathbf{x}) = \arg\min ||\mathcal{P}(\mathbf{x} + \mathbf{s}(\mathbf{x}))-\mathcal{P}(\mathbf{x}) ||_2^2
\end{equation}
\begin{equation}
	Res = \arg\min\sum\nolimits_{{\mathbf{x}\in\Omega}}|\mathcal{P}(\mathbf{x} + \mathbf{s}(\mathbf{x}))-\mathcal{P}(\mathbf{x})|
\end{equation}
\begin{equation}
	I_{enhanced} = I + \alpha \times f(Res)
\end{equation}

\section{Materials and methods}
The overall structure of the method is shown in \textcolor{cyan}{Fig .1}.  The method can be divided into three parts.  (i) First, the residual feature $ Res $ of the input image $ I $ is obtained by minimizing $ E(\rm{\mathbf{x}}) $. (ii) Second, a new feature refinement mechanism  $ f_{new}(*) $ is designed, through which the residual feature is updated. (iii) Third, the detail-enhanced image $ I_{enhanced} = I + \alpha \times f_{new}(Res) $ is acquired. Here, $ \alpha $ is a parameter that needs to be manually adjusted, which is used to ensure that $ I_{enhanced} $ can achieve a better visual effect.
\begin{figure}[h]	
	\centering
	\includegraphics[scale=1]{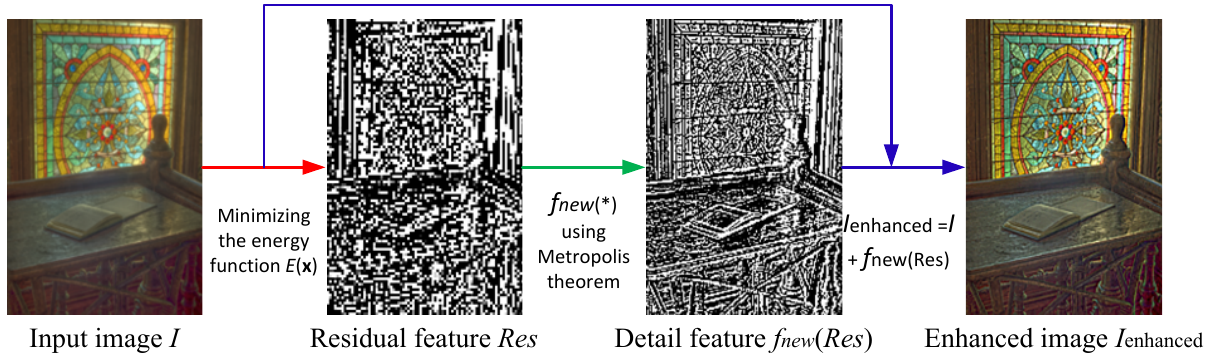}
	\caption{The architecture of the detail enhancement system based on Metropolis Theorem.}
\end{figure}

\subsection{Residual feature extraction}
As \textcolor{cyan}{Fig .1} shows, the process of initializing the residual feature is different from the method in [\textcolor{cyan}{5}, \textcolor{cyan}{6}],  this difference reflects in two aspects. Firstly, only local features are used in [\textcolor{cyan}{5}, \textcolor{cyan}{6}], but the features exploited in our method are almost features from non-local regions. Secondly, the energy function is different, only pixel loss $ E_{pixel}(\mathbf{x}) $ is considered in
[\textcolor{cyan}{5}, \textcolor{cyan}{6}], as Eq. (7) shows, but the energy function in our proposed method is redesigned, two new regularization terms force the matching process to focus not only on the pixel differences but on edge sharpness and texture smoothness as well, namely $ E_{gradient}(\mathbf{x}) $ and $ E_{smooth}(\mathbf{x}) $ defined in Eq. 8 and Eq. 9. Energy function $ E(\mathbf{x}) $ is a new patch matching criterion, and it is optimized to get $ \mathbf{s}(\mathbf{x}) $, i.e. $ \mathbf{s}(\mathbf{x}) = arg{\rm min} E(\mathbf{x}) $.    $\nabla(\cdot)$ and $ \nabla^2(\cdot) $ denote the Hamilton and Laplacian operators for a certain patch $ \mathcal{P}(\mathbf{x}) $, respectively, $ \Omega $ is the feasible region of $ \mathbf{x} $, and $ \eta $ and $ \mu $ represent the positive regularization constants to control the contributions of the prior components.

\begin{equation}
	E(\mathbf{x}) = E(\mathbf{x}) + \eta E_{gradient}(\mathbf{x}) + \mu E_{smooth}(\mathbf{x})
\end{equation}
\begin{equation}
	E_{pixel}(\mathbf{x}) = \sum\nolimits_{{\mathbf{x}\in\Omega}}||\mathcal{P}(\mathbf{x} + \mathbf{s}(\mathbf{x}))-\mathcal{P}(\mathbf{x}) ||_2^2
\end{equation}
\begin{equation}
	E_{gradient}(\mathbf{x}) = \sum\nolimits_{{\mathbf{x}\in\Omega}}||\nabla\mathcal{P}(\mathbf{x} + \mathbf{s}(\mathbf{x}))-\nabla\mathcal{P}(\mathbf{x}) ||_2^2
\end{equation}
\begin{equation}
	E_{smooth}(\mathbf{x}) = \sum\nolimits_{{\mathbf{x}\in\Omega}}||\nabla^2\mathcal{P}(\mathbf{x} + \mathbf{s}(\mathbf{x}))-\nabla^2\mathcal{P}(\mathbf{x}) ||_2^2
\end{equation}

As can be seen in \textcolor{cyan}{Fig .2}, the topology diagrams of six residual feature extraction methods are shown, among which three algorithms are based on deep learning techniques, and they obtain the initial residual features through different network structures, and then update the features by different losses, i.e. $ L_1 $, $ L_2 $ or perceptual loss. Different from these approaches, algorithms based on statistical learning, especially those based on residual learning, have their own unique designs on the residual feature updations. For example, ZF [\textcolor{cyan}{9}] continuously restores the details of the original images through zero order reverse filter, IPRH [\textcolor{cyan}{5}, \textcolor{cyan}{6}] updates the residual layer through fine searching and matching. Our proposed algorithm compares the whole system to a thermodynamic system and updates the residual features in a physical-based way.

\begin{figure}[h]	
	\centering
	\includegraphics[scale=1.3]{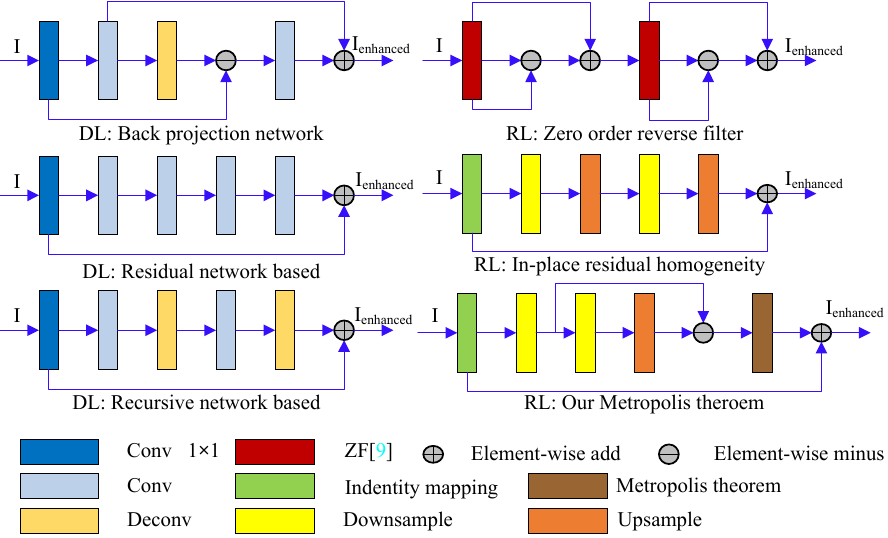}
	\caption{Topology diagrams of six algorithms for residual features extraction. The three algorithms on the left are based on Deep Learning (DL), and the other three on the right are based on Residual Learning (RL). Different modules are color-coded and annotated accordingly.}
\end{figure}

\subsection{Metropolis theorem}
As \textcolor{cyan}{Fig .3} shows, in the process of looking for the best matching patches, the energy function $ E(\mathbf{x}) $ in Eq. 6  converges in two local optimal states A and B, but what we really need is the global minimum state C. In fact, with the exception of meaningless full searching, almost all types of searching methods may fall into the local optimal
state. This is because searching is a greedy algorithm, that is, the energy function can only accept a better state (lower-energy point in \textcolor{cyan}{Fig .3}) than the current state in the iterative process, which makes the system very easy to fall into the local optimal state. For example, it’s easy for energy function $ {E}(\mathbf{x}) $ to go from state 1 to state B, but it’s hard for it to go from state
B to state 2 and then to final state C.

\begin{figure}[h]	
	\centering
	\includegraphics[scale=1]{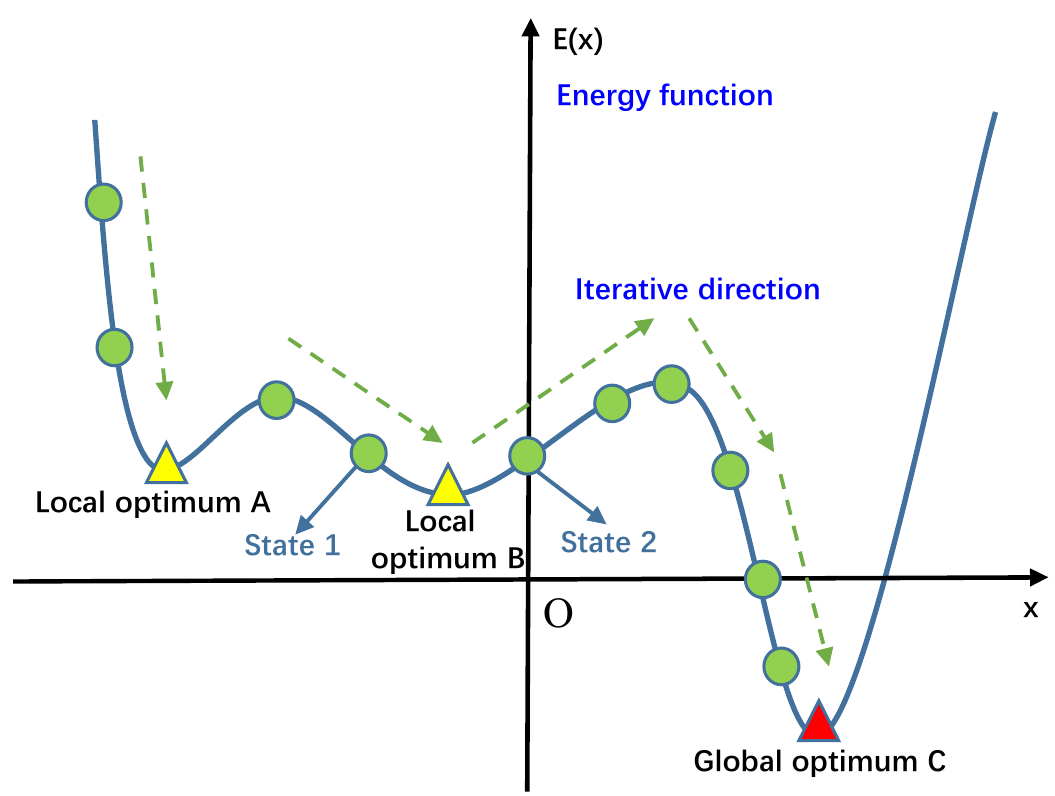}
	\caption{Schematic diagram of the principle of finding the lowest point of internal energy of a system using the laws of thermodynamics, also known as Metropolis theorem.}
\end{figure}

In our proposed method, the updating mechanism, known as $ f_{new}(\cdot) $, is redesigned. To find the global optimal solution, the updating mechanism is analogous to the cooling process of a thermodynamic system. In the thermodynamic system, the Metropolis theorem can find the lowest internal energy point of the whole system and this lowest internal energy point is the global optimal solution of our system, where a more suitable feature can be found for image detail enhancement. Therefore, we adopt the idea of Metropolis theorem [\textcolor{cyan}{32}] in thermodynamics. The core idea of this theorem is that if the next state is better than the current state, i.e. $ E_{n+1}(\mathbf{x})\leq E_{n}(\mathbf{x}) $, we accept the next state, but if the next state is worse than the current state, i.e. $ E_{n+1}(\mathbf{x})\textgreater E_{n}(\mathbf{x}) $, we still accept the next state with a probability $ p $. In Eq. 10, $ E\mathcal{L}(\mathbf{x}) $ is the energy loss related to patch $ \mathbf{x} $, and $ E\mathcal{L}(\mathbf{x}) = E_{n+1}(\mathbf{x})- E_{n}(\mathbf{x}) $, Boltzmann constant $ k = 1.38\times 10^{-23} $, and $ T $ means the temperature of the thermodynamic system. In practice, $ T_n $ is used 
instead of $ kT $ to describe the temperature of the current state. To get the most stable state, namely the point called global optimum C, the system’s temperature is decreased in each iteration by setting $ T_{n} \leftarrow \gamma \times T_{n} $, and the physical meaning of $ \gamma $ is the cooling coefficient of the thermodynamic system.
\begin{equation}
	 p = {\exp}(-\frac{E\mathcal{L}(\mathbf{x})}{kT})
\end{equation}

For the convenience of hardware implementation, the mathematical equivalent of infinitesimal property is used to rewrite the probability expression, i.e. $ {\exp}(-\frac{E\mathcal{L}(\mathbf{x})}{kT}) \approx 1-\frac{E_{n+1}(\mathbf{x})- E_{n}(\mathbf{x})}{T_{n}} $, as Eq.11 shows. The hardware is slow
for exponential operation,  but fast for addition, subtraction, and multiplication. This improvement caters to the hardware architecture and is conducive to improving the overall speed of the program.
\begin{equation}
\begin{aligned}
	\begin{split}
		p= \left \{
		     \begin{array}{lr}
			1    & E_{n+1}(\mathbf{x})\leq E_{n}(\mathbf{x})\\
			1-\frac{E_{n+1}(\mathbf{x})- E_{n}(\mathbf{x})}{T_{n}} & E_{n+1}(\mathbf{x})\textgreater E_{n}(\mathbf{x})
		\end{array}\right.
		\end{split}	
\end{aligned}
\end{equation}

To prove the validity of the Metropolis theorem, in the experiment, the image patch on the left is the original patch to be matched. The middle image patch is with the smallest pixel difference from the left patch using the algorithm in [\textcolor{cyan}{5}, \textcolor{cyan}{6}]. The image patch on the right is the closest structure to the left patch by using the Metropolis theorem. As can be seen from \textcolor{cyan}{Fig. 4}, the best-matching pair found by the method in [\textcolor{cyan}{5}, \textcolor{cyan}{6}] is relatively coarse, and it isn’t structured similarly to the original patch. With the help of the new matching criterion and updating mechanism, the system takes into account both the gradient and the texture information and finds an image patch that is structurally more similar to the original patch.

\begin{figure}[h]	
	\centering
	\includegraphics[scale=0.6]{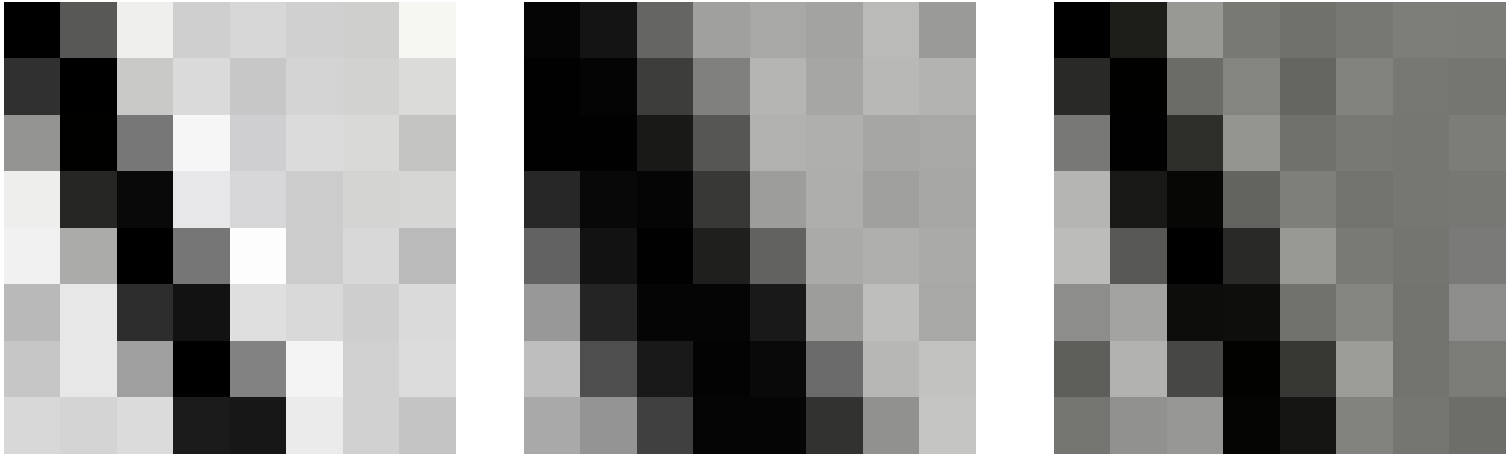}
	\caption{From left to right: One original patch, and the best-matching pairs found by method in [\textcolor{cyan}{5}, \textcolor{cyan}{6}]
		and our method respectively.}
\end{figure}

\subsection{In-place matching mechanism}
The algorithm in [\textcolor{cyan}{5}, \textcolor{cyan}{6}] can be simply summarized as  $ \mathbf{s}(\mathbf{x}) = [T_x, T_y] = \arg\min\sum_{T_x = -2}^{2}\sum_{T_y = -2}^{2}||\mathcal{P}(\mathbf{x} + \mathbf{s}(\mathbf{x}))-\mathcal{P}(\mathbf{x})||_2^2 $. Later, the in-place matching mechanism is proposed in case of only using four adjacent pixels, i.e. $ \mathbf{s}(\mathbf{x})' = [T_x', T_y'] = \arg\min\sum_{T_x' = 0}^{1}\sum_{T_y' = 0}^{1}||\mathcal{P}(\mathbf{x} + \mathbf{s}(\mathbf{x}))-\mathcal{P}(\mathbf{x})||_2^2 $, though which the system can suffer only a negligible PSNR loss with an about 90\% reduction in runtime. As mentioned above, this approach only considers the pixel loss and ignores the texture structure of an image, which is very important for a detail enhancement algorithm. Therefore, the matching method of the algorithm is rewritten to guarantee its matching efficiency and force it to focus on the edge and texture structure features as well, as Eq. 12 shows.

\begin{equation}
	\mathbf{s}(\mathbf{x})'' = [T_x'', T_y''] = \arg\min\sum_{T_x'' = 0}^{1}\sum_{T_y'' = 0}^{1} \\||E(\mathbf{x} + \mathbf{s}(\mathbf{x}))-E(\mathbf{x})||_2^2
\end{equation}

Note that the feasible region of $ \mathbf{x} $ in Eq. 12 has both local features in in-place regions and non-local features generated by the Metropolis theorem, which not only makes the system more likely to converge to the global optimal solution, but also makes the sparsity of the feature space more obvious, and the sparsity property is more favorable to the generation of the image detail layer. 

\begin{algorithm}[h] 
	\caption{Searching and matching mechanism based on Metropolis theorem}
	\textbf{Input:} The initial temperature  $ T_0$,  the cooling coefficient $ \gamma $, number of loop execution $ n $, number of seeds used for diffusion $ N $.\\
	\textbf{Output:} The best matching image patch $ v $ of the original image patch $ u $.\\
	1. \textbf{Repeated:}\\
	2.\quad\quad For each $ u $,  Eq.12 is used to obtain the to-be-matched patches $ \{v_1,\cdots v_N\} $.\\
	3.\quad\quad Compute $ E(u) $ and $ E(v_i) $ using Eq. 6. $ v_* = \arg\min\nolimits_{i\in \{1\cdots N\}} |E(u)-E(v_i)| $\\
	4.\quad\quad Compute energy loss $ E\mathcal{L}_i= |E(u)-E(v_i)|-|E(u)-E(v_*)| $,  $ i\in \{1\cdots N\} $\\
	5.\quad\quad Generate a random number $ K $ within $ (0,1) $.\\
	6.\quad\quad Eq.11 is used to compute $ p $, \textbf{If} ($ p > K $) \quad  $ v_* \leftarrow v_i \bigcup{v_*} $ \quad \textbf{End If} \\
	7.\quad\quad Let $ v_* $ be $ u $, $ n \leftarrow n-1 $, update $ T_i $ through $ T_i \leftarrow \gamma \times T_i $\\
	8. \textbf{Until:} $ n<0 $  or $ T_i< 0.001 $\\
	9. In the last loop, for all of the $ v_* $, $ v = \arg\min |E(u)-E(v_*)|$\\
	\textbf{Return:} The best image matching patch $ v $. 
\end{algorithm}

\section{Experimental analysis}
\subsection{Datasets}
Six datasets are used in this experiment, including four natural image datasets and two medical image datasets. The four natural image datasets are Set5 [\textcolor{cyan}{15}], Set14 [\textcolor{cyan}{16}],
BSD100 [\textcolor{cyan}{17}] and General100 [\textcolor{cyan}{34}]. These four datasets are used internationally as texture datasets. Among them, Set5 [\textcolor{cyan}{15}] and Set14 [\textcolor{cyan}{16}] contain complex scenes, BSD100 [\textcolor{cyan}{17}] contains images with rich frequencies, and General100 [\textcolor{cyan}{34}] contains images with very complicated structures.  The other two are publicly available medical datasets named CVC and EITS. These two datasets contain more than 6000 unlabeled and 400 annotated images respectively, which are authorized to be used for free in academic research.

\subsection{Experimental setting}
The running platform of this experiment is Matlab 2014b. In this paper, many effective algorithms are used for comparisons. They are the local filter-based algorithms GIF [\textcolor{cyan}{2}], RGIF [\textcolor{cyan}{7}], GGIF [\textcolor{cyan}{3}], EGIF [\textcolor{cyan}{8}], SPGIF [\textcolor{cyan}{15}], SGIF [\textcolor{cyan}{31}], the global filter-based algorithms WLS [\textcolor{cyan}{11}], FS [\textcolor{cyan}{13}], BFLS [\textcolor{cyan}{21}], ILS [\textcolor{cyan}{10}], TH [\textcolor{cyan}{33}], and the residual learning-based method ZF [\textcolor{cyan}{9}], LSE [\textcolor{cyan}{27}], IPRH [\textcolor{cyan}{6}], respectively. 
All the codes associated with this paper are licensed and can be downloaded free from Github, and their default parameters are followed.

In our algorithm, the experimental parameters are set as follows: cooling coefficient $ \gamma = 0.98 $, initial thermodynamic temperature $ T_0 = 300 $, and the system execution number $ n = 20 $. The above parameter settings not only refer to the physical meaning in the thermodynamics system but are determined in the number of experimental tests as well.

\subsection{Objective metrics comparisons}
The objective evaluation metrics used in this paper are RMSE (Root Mean Square Error) and SSIM [\textcolor{cyan}{18}] (Structural SIMilarity), both of which are internationally recognized and used in the field of image quality evaluation. RMSE is a measure of the difference between image pixel domains, and a smaller value means that the corresponding algorithm is better. SSIM is a measure of the structural similarity between two images in the range of 0 to 1, and the closer the value of SSIM is to 1, the more effective the corresponding algorithm is. All quantization tests are performed on the Y channel of the image. 

\begin{equation}
	{\rm{RMSE}} = \sqrt{\frac{1}{m\times n}\sum_{i=1}^m\sum_{j=1}^n(I_{gt}(i,j)-I'(i,j))^2}
\end{equation}
\begin{equation}
	{\rm{SSIM}} = \frac{(2\mu_x\mu_y+c_1)(2\delta_{xy}+c_2)}{(\mu_x^2+\mu_y^2+c_1)(\delta_x^2+\delta_y^2+c_2)}
\end{equation}

The RMSE and SSIM are calculated as shown in Eq. 13 and Eq. 14, where $  m\times n $ is the resolution of images $ I_{gt} $ and $ I' $, $ I_{gt} $ is the ground truth image, and $ I' $ is the result of a specific algorithm. $ \mu_x $, $ \mu_y $ and $ \delta_x^2 $, $ \delta_y^2 $ are the mean and variance of $ I_{gt} $ and $ I' $ respectively, $ \delta_{xy} $ is the covariance of $ I_{gt} $ and $ I' $, and $ c_1 $ and $ c_2 $ are very small constants and their values are set to 0.001 in the paper.

\begin{table}[htbp]	
	\caption{Quantitative tests.  Average RMSE/SSIM for detail layer magnification factor $ \alpha $ $ \times $ 2,  $ \times $4 on Set5[\textcolor{cyan}{26}], Set14[\textcolor{cyan}{27}], BSD100[\textcolor{cyan}{28}], and General100[\textcolor{cyan}{34}] datasets of our method based on Metropolis Therorem (MT) and other methods. The best and second best results are shown in \textbf{black bold} and \textbf{\textcolor{blue}{blue bold}}.}
	\begin{small}
		\setlength{\tabcolsep}{1.5mm}{
			\begin{tabular}{cccccc}  
				\hline
				Method & Factor  & Set5[\textcolor{cyan}{26}]   & Set14[\textcolor{cyan}{27}]   & BSD100 [\textcolor{cyan}{28}]   & General100[\textcolor{cyan}{34}]    \\\hline
				GIF[\textcolor{cyan}{1,2}] &   $ \times $ 2    & 4.46/0.9960 & \textbf{\textcolor{blue}{4.54}}/0.9912 & \textbf{4.06}/0.9946 & \textbf{\textcolor{blue}{4.18}}/0.9952 \\ 
				RGF[\textcolor{cyan}{7}]      &   $ \times $ 2    & 8.07/0.9966 & 11.52/0.9941 & 16.65/\textbf{\textcolor{blue}{0.9958}} &     5.20/0.9987    \\
				GGIF[\textcolor{cyan}{3}]   &   $ \times $ 2    & 12.06/0.9723 & 13.14/0.9254 & 28.25/0.8139 &  14.12/0.9315  \\
				EGIF[\textcolor{cyan}{8}]    &   $ \times $ 2    & 10.75/0.9786 & 12.30/0.9514 & 27.85/0.8266 &  12.14/0.9621 \\
				SPGIF[\textcolor{cyan}{14}]  &   $ \times $ 2    & 22.92/0.9528 & 17.97/0.9474 & 8.50/0.9982 &  14.91/0.9820 \\
				WLS[\textcolor{cyan}{11}]  &   $ \times $ 2    & 22.69/0.9200 & 26.39/0.8131 & 21.26/0.9116 &  27.23/0.8539\\
				FS[\textcolor{cyan}{13}]    &   $ \times $ 2   & \textbf{\textcolor{blue}{2.37}}/0.9983 & 4.55/0.9963 & \textbf{\textcolor{blue}{8.12}}/0.9949 &  2.26/0.9986\\
				BFLS[\textcolor{cyan}{21}]    &   $ \times $ 2    & 15.60/0.9502 & 11.04/0.9495 & 13.21/0.9351 &  13.92/0.9364\\  
				ILS[\textcolor{cyan}{10}]  &   $ \times $ 2    & 16.29/0.9661 & 10.89/0.9728 & 11.36/0.9745 & 16.04/0.9551\\
				TH[\textcolor{cyan}{33}]    &   $ \times $ 2   & 10.65/0.9838 & 11.32/0.9816 & 21.95/0.9542 & 7.50/0.9903\\
				ZF[\textcolor{cyan}{9}]    &   $ \times $ 2    & 11.07/\textbf{\textcolor{blue}{0.9989}} & 18.28/0.9964 & 29.12/0.9915 & 8.59/\textbf{\textcolor{blue}{0.9993}} \\
				IPRH[\textcolor{cyan}{5,6}] &   $ \times $ 2    & 3.74/0.9988 & 5.46/\textbf{\textcolor{blue}{0.9988}} & 12.62/0.9939 & 5.20/0.9972 \\
				Our MT &   $ \times $ 2    & \textbf{1.64}/\textbf{0.9999} & \textbf{4.18}/\textbf{0.9991} & 8.20/\textbf{0.9973} & \textbf{1.68}/\textbf{0.9998} \\
				\hline	
				GIF[\textcolor{cyan}{1,2}] &   $ \times $ 4    & 15.04/0.9595 & 11.24/0.9549 & 10.61/0.9825  & 13.85/0.9855 \\ 
				RGF[\textcolor{cyan}{7}]      &   $ \times $ 4    & 24.49/0.9916 & 17.66/0.9841 & 14.49/0.9899  & 18.06/0.9960        \\
				GGIF[\textcolor{cyan}{3}]   &   $ \times $ 4    & 21.27/0.8683 & 18.27/0.8822 &  12.46/0.9720 & 24.32/0.9455\\ 
				EGIF[\textcolor{cyan}{8}]    &   $ \times $ 4    & -/- & -/- &  -/- & -/- \\
				SPGIF[\textcolor{cyan}{14}]  &   $ \times $ 4    & 67.19/0.5748 & 48.67/0.5938 & 92.19/0.4379  & 67.82/0.7365 \\
				WLS[\textcolor{cyan}{11}]    &   $ \times $ 4    & -/- & -/- & -/-  & -/- \\
				FS[\textcolor{cyan}{13}]    &   $ \times $ 4    & \textbf{5.94}/0.9892 & \textbf{4.54}/0.9900 & \textbf{3.59}/0.9904  & \textbf{3.43}/\textbf{\textcolor{blue}{0.9974}} \\
				BFLS[\textcolor{cyan}{21}]     &   $ \times $ 4    & 20.67/0.8900 & 16.36/0.9121 & 13.84/0.9661 & 25.55/0.9378 \\
				ILS[\textcolor{cyan}{10}]   &   $ \times $ 4   & 21.68/0.9357 & 17.31/0.9447 & 33.19/0.9265 & 27.57/0.9680 \\
				TH[\textcolor{cyan}{33}]    &   $ \times $ 4    & 24.84/0.9581 & 16.97/0.9556 & 14.01/0.9741 & 25.31/0.9644 \\  
				ZF[\textcolor{cyan}{9}]    &   $ \times $ 4    & 35.15/0.9836 & 31.78/0.9646& 23.57/0.9725 & 29.06/0.9983 \\
				IPRH[\textcolor{cyan}{5,6}] &   $ \times $ 4    & 11.08/\textbf{\textcolor{blue}{0.9987}} & 9.48/\textbf{\textcolor{blue}{0.9961}} & 9.89/\textbf{\textcolor{blue}{0.9943}} & 9.47/0.9971 \\
				Our MT &   $ \times $ 4    & \textbf{\textcolor{blue}{7.63}}/\textbf{0.9994} & \textbf{\textcolor{blue}{7.46}}/\textbf{0.9973} & \textbf{\textcolor{blue}{4.82}}/\textbf{0.9996} & \textbf{\textcolor{blue}{4.98}}/\textbf{0.9996} \\
				\hline		
		\end{tabular}}	
	\end{small}	
\end{table}

The following conclusions can be drawn from the statistics in \textcolor{cyan}{Table 1} and \textcolor{cyan}{Table 2}. First, our algorithm achieves the best results in all SSIM tests, which shows that our algorithm has the strongest ability to protect the structural information of the image while doing the detail enhancement task. Second, our algorithm also achieves the first and second positions in most of the RMSE tests, which confirms the ability of our algorithm to protect the pixel domain information while detail enhancement. Third, our RMSE ranking gradually slips to the second position as the factor increases, which indicates that our proposed algorithm prioritizes the structural information of the image during detail enhancement, resulting in a better visual performance of the final outputs. It should be noted that the detail amplification factors in WLS[\textcolor{cyan}{11}] and EGIF[\textcolor{cyan}{8}] are globally generated and adaptive, so there is no amplification factor equal to 4 for these two algorithms.

\begin{table}[htbp]	
	\caption{Quantitative tests.  Average RMSE/SSIM for detail layer magnification factor $ \alpha $ $ \times $ 2,  $ \times $4 on CVC and EITS datasets of our method based on Metropolis Theorem (MT) and other methods. The best and second best results are shown in \textbf{black bold} and \textbf{\textcolor{blue}{blue bold}}.}
	\begin{small}
		\setlength{\tabcolsep}{2.9mm}{
			\begin{tabular}{ccccc}  
				\hline
				Method   & CVC/ $ \times $ 2   & CVC/ $ \times $ 4   & EITS/ $ \times $ 2   & EITS/ $ \times $ 4    \\\hline
				GIF[\textcolor{cyan}{1,2}]     & 2.78/0.9958 & 7.73/0.9820 & 3.56/0.9952 & 9.75/0.9665 \\ 
				RGF[\textcolor{cyan}{7}]          & 3.75/\textbf{\textcolor{blue}{0.9993}} & 7.37/0.9973 & 6.26/\textbf{\textcolor{blue}{0.9985}} &     11.42/0.9947    \\
				GGIF[\textcolor{cyan}{3}]       & 4.73/0.9791 & 8.73/0.9646 & 8.47/0.9474 &  13.90/0.9025  \\
				EGIF[\textcolor{cyan}{8}]       & 3.44/0.9977 & -/- & 5.75/0.8266 &  -/- \\
				SPGIF[\textcolor{cyan}{14}]      & 31.80/0.8937 & 72.30/0.4947 & 17.37/0.9652 &  51.43/0.6625 \\
				WLS[\textcolor{cyan}{11}]      & 20.43/0.9161 & -/- & 23.46/0.8757 &  -/-\\
				FS[\textcolor{cyan}{13}]       & \textbf{\textcolor{blue}{1.21}}/0.9989 & \textbf{1.21}/0.9974 & \textbf{\textcolor{blue}{2.42}}/0.9983 &  \textbf{2.42}/\textbf{\textcolor{blue}{0.9959}}\\
				BFLS[\textcolor{cyan}{21}]       & 6.29/0.9665 & 10.82/0.9447 & 10.36/0.9705 &  17.82/0.9254\\  
				ILS[\textcolor{cyan}{10}]     & 11.82/0.9779 & 21.05/0.9265 & 13.62/0.9644 & 22.74/0.9102\\
				TH[\textcolor{cyan}{33}]        & 4.97/0.9977 & 9.34/0.9924 & 6.06/0.9956 & 11.22/0.9858\\
				ZF[\textcolor{cyan}{9}]       & 6.93/0.9973 & 18.52/0.9809 & 12.44/0.9954 & 24.38/0.9809 \\
				IPRH[\textcolor{cyan}{5,6}]    & 1.55/0.9990 & 2.94/\textbf{\textcolor{blue}{0.9974}} & 3.63/0.9973 & 6.91/0.9925 \\
				Our MT     & \textbf{1.13}/\textbf{0.9997} & \textbf{\textcolor{blue}{2.16}}/\textbf{0.9990} & \textbf{1.53}/\textbf{0.9999} & \textbf{\textcolor{blue}{2.85}}/\textbf{0.9995} \\
				\hline		
		\end{tabular}}	
	\end{small}	
\end{table}

\subsection{Visual performance}
In daily life, a large amount of information is obtained through human eyes. Since the human eye is an intelligent discriminator that can effectively judge the effectiveness of visual tasks, the visual performance comparisons of images are particularly important.
In this paper, the comparison of the subjective effects of the images and the ranking of MOS metrics are used to highlight the subjective visual superiority of our algorithm.

\textcolor{cyan}{Fig. 5} $ \sim $ \textcolor{cyan}{Fig. 9} show some detail-enhanced result images, from which some conclusions can be drawn. Firstly, the results of some algorithms are visually poorly experienced, most notably the results of SPGIF [\textcolor{cyan}{14}] and ZF[\textcolor{cyan}{9}]. SPGIF [\textcolor{cyan}{14}] has successful applications for medical images, but for natural images, it often changes their chromaticity. In addition, when the detail magnification factor becomes larger, the ZF [\textcolor{cyan}{9}] results have a large amount of white noise attached to the texture of the image, which makes the texture no longer clear. Secondly, some algorithms change the inherent color of things in the input image while detail enhancement, which is reflected in the results of algorithms RGIF [\textcolor{cyan}{7}], GGIF [\textcolor{cyan}{3}], BFLS [\textcolor{cyan}{21}], ILS [\textcolor{cyan}{10}], and TH [\textcolor{cyan}{33}]. For example, in \textcolor{cyan}{Fig. 6}, the GGIF [\textcolor{cyan}{3}], BFLS [\textcolor{cyan}{21}], and ILS [\textcolor{cyan}{10}] change the color of the goldfish's eyes from white to green, in \textcolor{cyan}{Fig. 7}, GIF [\textcolor{cyan}{1, 2}], GGIF [\textcolor{cyan}{3}], BFLS [\textcolor{cyan}{21}], TH [\textcolor{cyan}{33}] changes the color of the dragonfly's legs from black to purple; in \textcolor{cyan}{Fig. 8}, the color of the clouds is also destroyed after the GGIF [\textcolor{cyan}{3}], BFLS [\textcolor{cyan}{21}], and ILS [\textcolor{cyan}{10}] processing. Thirdly, some algorithms can damage the texture of the image. In \textcolor{cyan}{Fig. 6}, GGIF [\textcolor{cyan}{3}], BFLS [\textcolor{cyan}{21}] and ILS[\textcolor{cyan}{10}] blacken the texture of water plants, in \textcolor{cyan}{Fig. 7}, ILS [\textcolor{cyan}{10}] distorts the texture of leaf veins, and the same phenomenon occurs in \textcolor{cyan}{Fig. 9}, where the texture of tree trunks is destroyed by the algorithms RGIF [\textcolor{cyan}{7}], BFLS [\textcolor{cyan}{21}] and TH [\textcolor{cyan}{33}].

\begin{figure}[htbp]	
	\centering
	\includegraphics[scale=0.68]{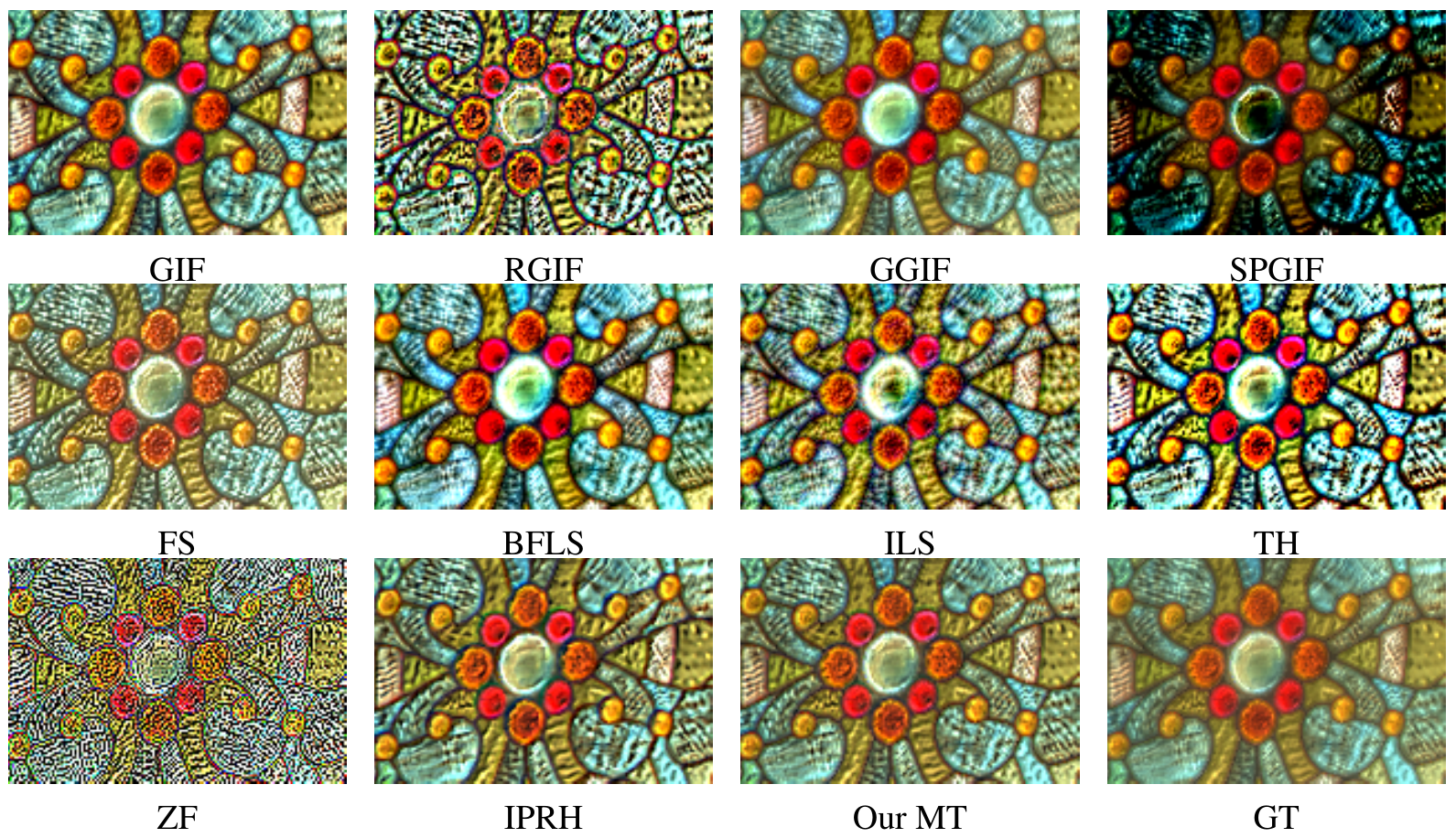}
	\caption{The first visual comparison. Each method's name is marked below the corresponding image. GT means the Ground Truth(input image) and detail layer magnification factor $ \alpha = 4 $. }
\end{figure}

\begin{figure}[htbp]	
	\centering
	\includegraphics[scale=0.75]{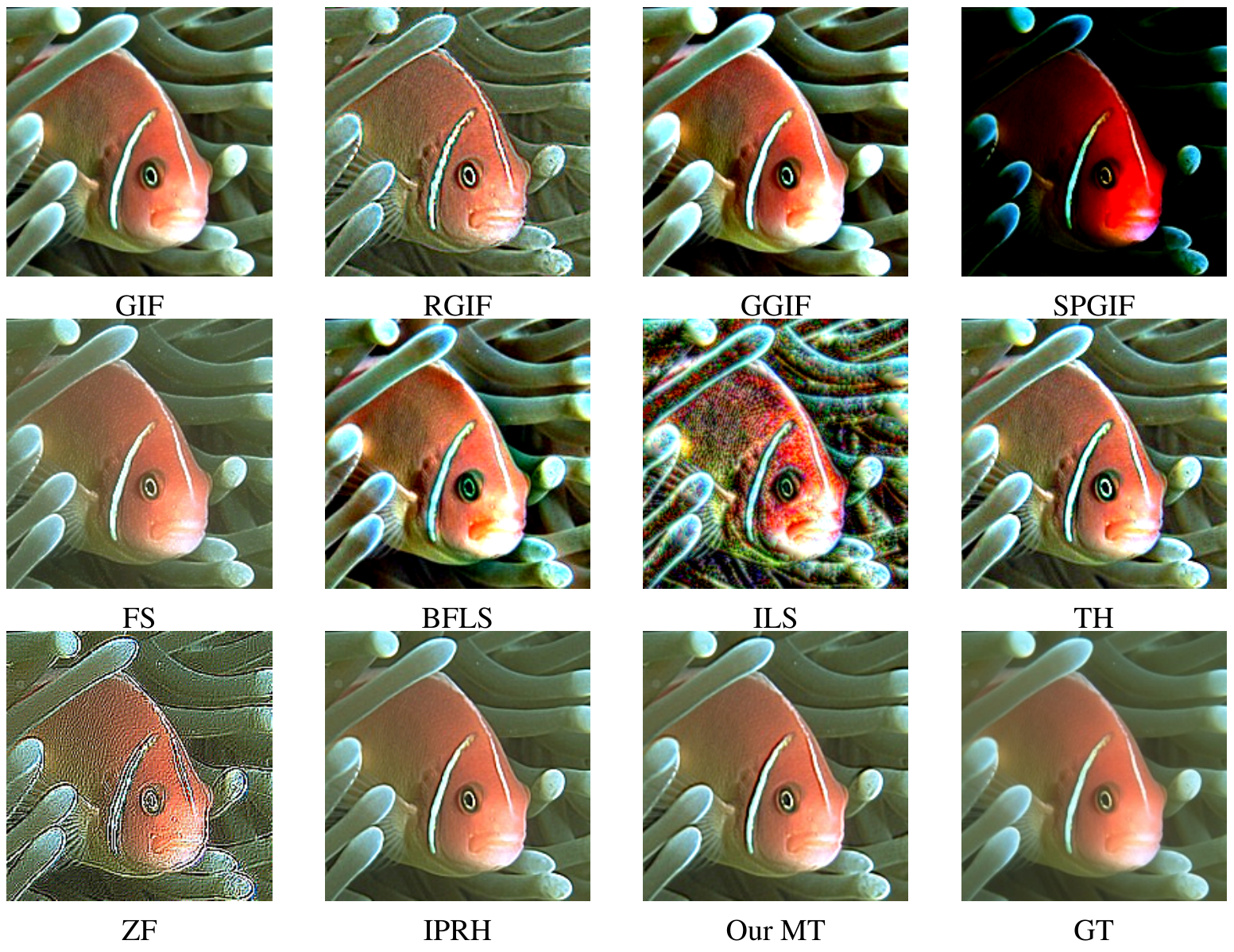}
	\caption{The second visual comparison. Each method's name is marked below the corresponding image. GT means the Ground Truth(input image) and detail layer magnification factor $ \alpha = 4 $.}
\end{figure}

\begin{figure}[htbp]	
	\centering
	\includegraphics[scale=0.75]{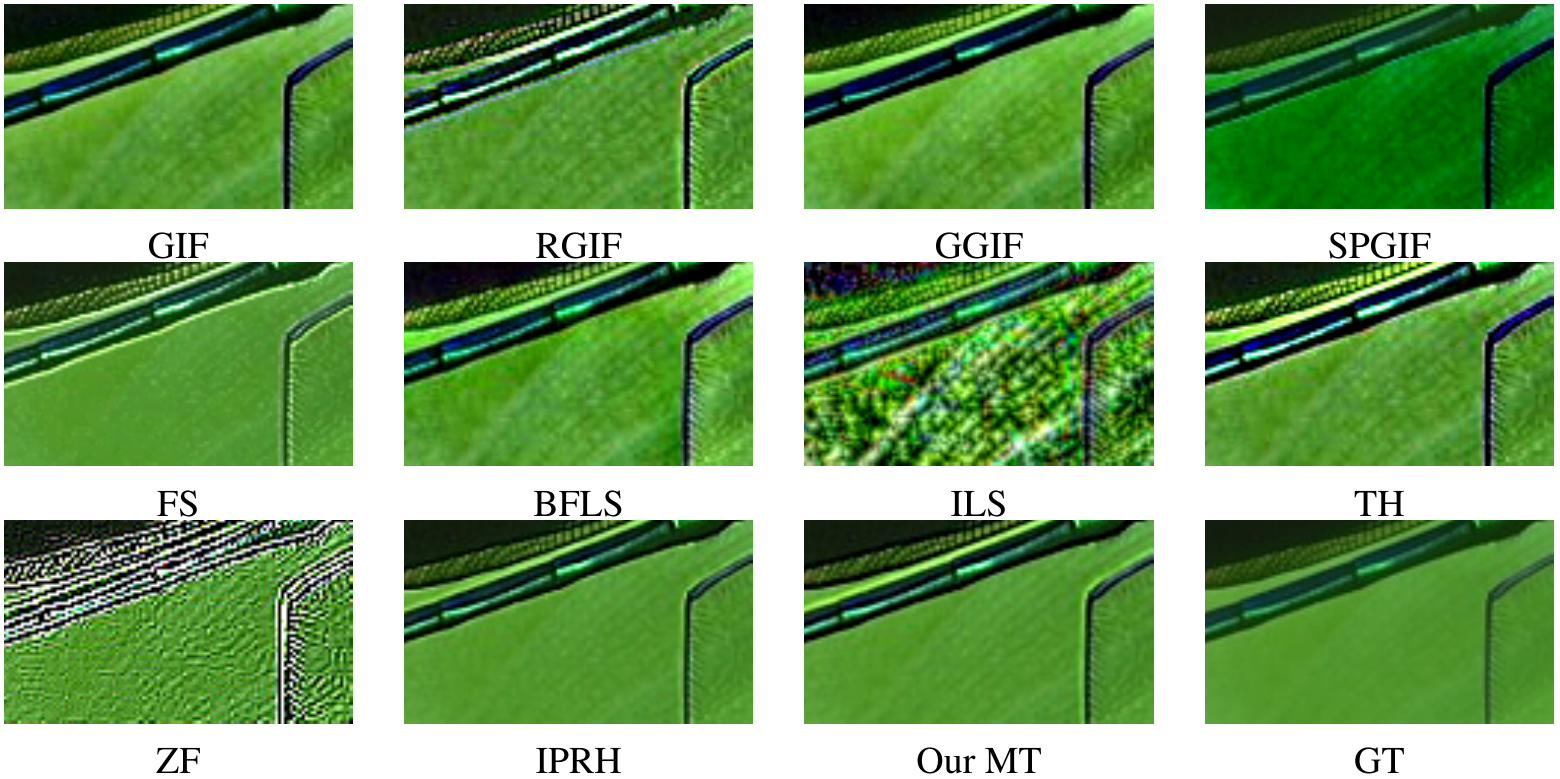}
	\caption{The third visual comparison. Each method's name is marked below the corresponding image. GT means the Ground Truth(input image) and detail layer magnification factor $ \alpha = 4 $.}
\end{figure}

\begin{figure}[htbp]	
	\centering
	\includegraphics[scale=0.75]{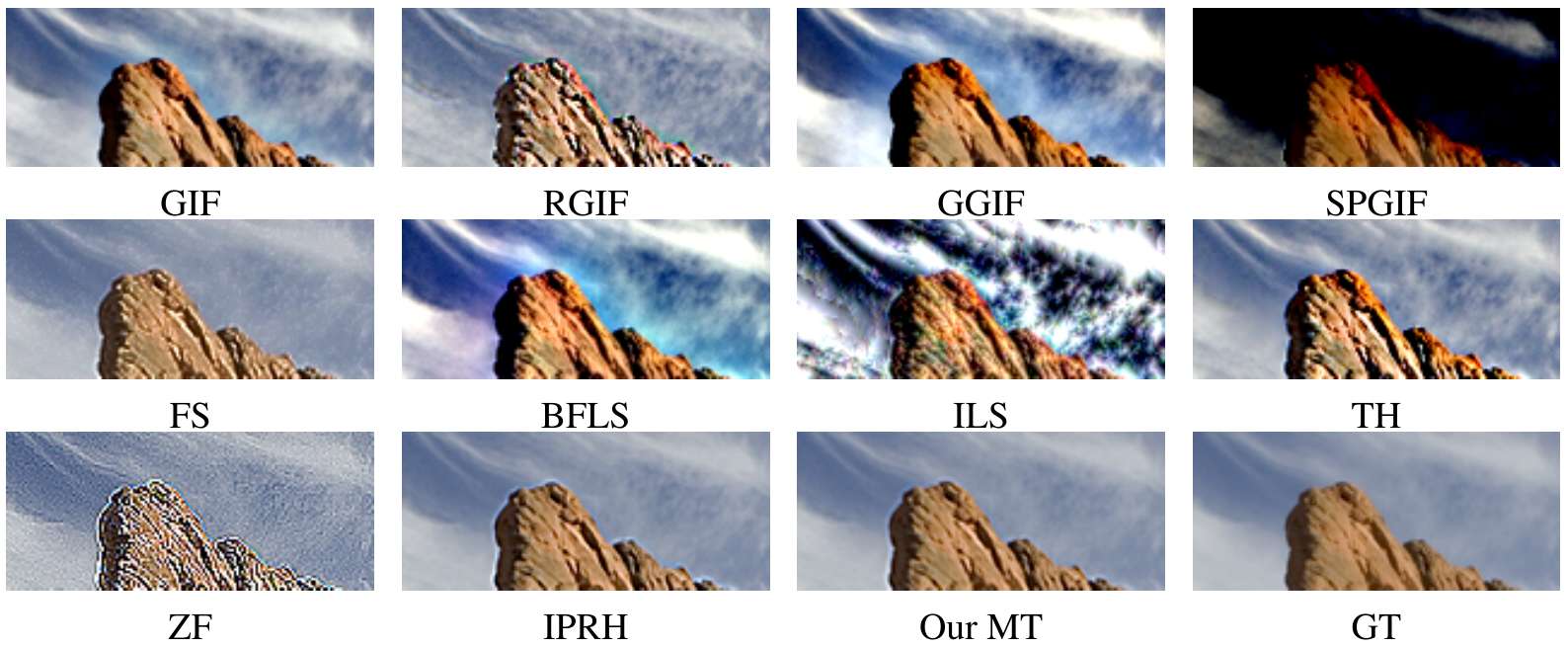}
	\caption{The fourth visual comparison. Each method's name is marked below the corresponding image. GT means the Ground Truth(input image) and detail layer magnification factor $ \alpha = 4 $.}
\end{figure}

\begin{figure}[htbp]	
	\centering
	\includegraphics[scale=0.75]{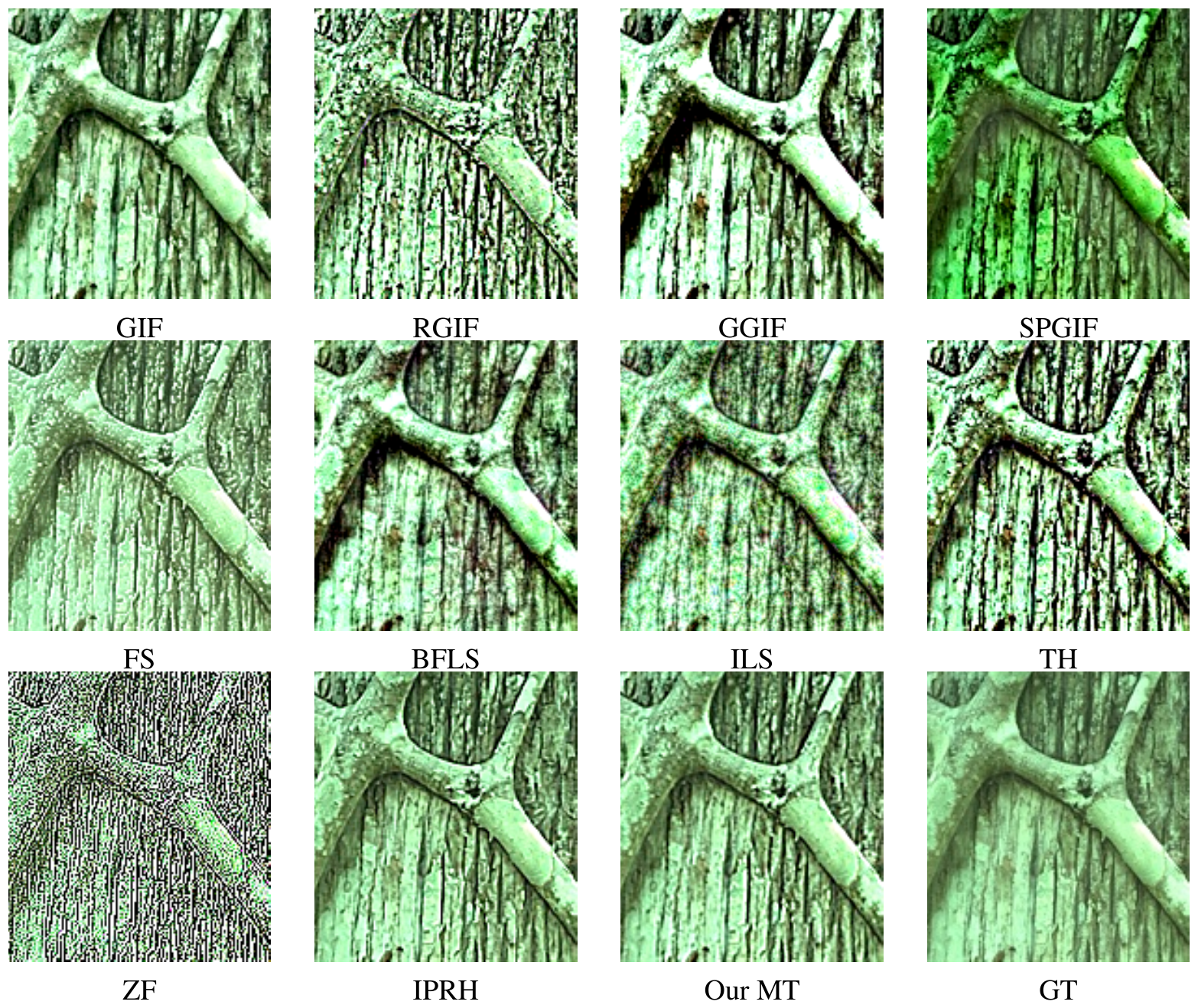}
	\caption{The fifth visual comparison. Each method's name is marked below the corresponding image. GT means the Ground Truth(input image) and detail layer magnification factor $ \alpha = 4 $.}
\end{figure}

In general, the FS [\textcolor{cyan}{13}], IPRH [\textcolor{cyan}{5, 6}], and MT algorithms are more effective, but there are some shortcomings in the FS [\textcolor{cyan}{13}] algorithm, for example, in \textcolor{cyan}{Fig. 6}, gradient reversal artifact exists on the dragonfly's abdomen after FS's [\textcolor{cyan}{13}] processing. In contrast, our algorithm is a texture-realistic detail enhancement algorithm, which enhances the original image while preserving the texture, structure, and other features of the image.

\subsection{MOS comparisons}
MOS is short for Mean Opinion Score, and it is a widely-used international subjective score evaluation metric of visual tasks. Specifically, it is to select several people with and without professional backgrounds according to the proportion and send the images to be evaluated to them for rating. The rating standard only depends on the comfort of human eyes during observation. Then the extreme scores are removed and the remaining scores are averaged in descending order to get the final result. 

\begin{table}[htbp]	
	\caption{MOS tests. Nine textures are selected and the top 5 score for each texture are shown.}
	\begin{small}
		\setlength{\tabcolsep}{7.9mm}{
			\begin{tabular}{cc}  
				\hline
				Textures  &   MOS ranking (Top 5) \\\hline
				Medical images  &   SPGIF $ \textgreater $ \textbf{MT} $ \textgreater $ GGIF $ \textgreater $ GIF $ \textgreater $ RGIF \\
				Clothes          &  \textbf{MT} $ \textgreater $ GIF $ \textgreater $ RGIF $ \textgreater $ WLS $ \textgreater $ GGIF\\
				Animal hides and skins  & TH $ \textgreater $ MT $ \textgreater $ GIF $ \textgreater $ BFLS $ \textgreater $ RGIF\\
				Anime portraits  & \textbf{MT} $ \textgreater $ RGIF $ \textgreater $ TH $ \textgreater $ GIF $ \textgreater $ BFLS\\
				Natural Landscape  &  \textbf{MT} $ \textgreater $ IPRH $ \textgreater $ FS $ \textgreater $ GIF $ \textgreater $ GGIF\\
				Printed posters    &   IPRH $ \textgreater $ \textbf{MT} $ \textgreater $ FS $ \textgreater $ GGIF $ \textgreater $ GIF\\
				Food \& Beverage           & \textbf{MT} $ \textgreater $ FS $ \textgreater $ IPRH $ \textgreater $ TH $ \textgreater $ RGIF\\
				Building \& Statues        & \textbf{MT} $ \textgreater $ FS $ \textgreater $ RGIF $ \textgreater $ IPRH $ \textgreater $ GIF \\
				Plants              &  \textbf{MT} $ \textgreater $ IPRH $ \textgreater $ BFLS $ \textgreater $ ILS $ \textgreater $ GIF     \\
				\hline		
		\end{tabular}}	
	\end{small}	
\end{table}

As can be seen in \textcolor{cyan}{Table 3}, our proposed detail enhancement algorithm, also known as MT, basically achieved first and second place in the MOS tests. This shows that our algorithm is a visually sound algorithm that satisfies the visual needs of most people. This also indirectly shows the robustness of our algorithm, which can effectively enhance a wide range of textures.

\subsection{Intensity curve analysis}
The intensity curve is a visualization method. Suppose there exists an image $ Y \in \mathcal{R}^{m\times n} $. A random row of pixels $ y \in \mathcal{R}^{1\times n} $ is taken from $ Y \in \mathcal{R}^{m\times n} $ for display, and the amplitude of this row of pixels reflects the variation of the image, and its variation property at the edges is an important property, i.e., the edge-preserving property. A signal with edge-preserving property can protect the gradient domain information more completely, thus making the image visually clearer after detail enhancement.

As can be seen in \textcolor{cyan}{Fig. 10}, during the detail enhancement process,  our algorithm fits this GT signal more closely than other algorithms in the region where the image amplitude changes abruptly, i.e., the gradient region, which indicates that our algorithm has a relatively stronger edge-preserving capability. This also indirectly supports why our proposed algorithm has a better detail enhancement effect.

\begin{figure}[h]	
	\centering
	\includegraphics[scale=0.45]{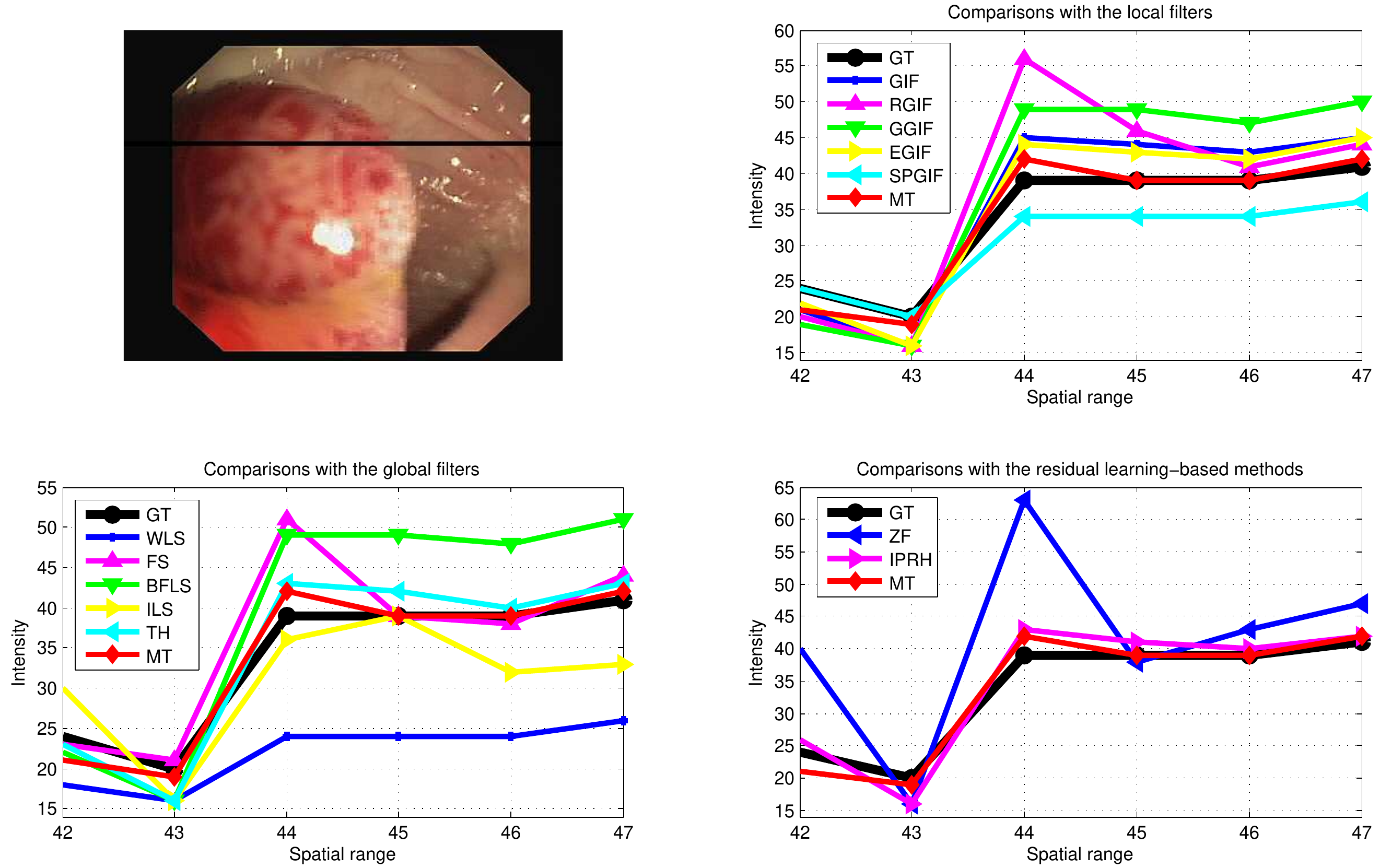}
	\caption{The architecture of the detail enhancement system based on Metropolis Theorem.}
\end{figure}

\subsection{Comparisons with deep learning-based algorithms}
Image enhancement algorithms based on deep learning include three types, low-illumination enhancement algorithm, blind enhancement algorithm, and detail enhancement algorithm. Low-illumination or blind enhancement methods usually change the image contrast or start from aesthetic-based models to improve the images' visual expression, i.e. methods in [\textcolor{cyan}{28}, \textcolor{cyan}{29}, \textcolor{cyan}{35}], and these two types algorithms have nothing to do with the algorithm discussed in this article. 

Due to their outstanding fitting and generalization capabilities, deep networks have excellent performance on many vision tasks. However, when it comes to detail enhancement, the results produced by deep networks are often unsatisfactory. Recently, many deep learning-based algorithms [\textcolor{cyan}{36}, \textcolor{cyan}{37}, \textcolor{cyan}{38}, \textcolor{cyan}{39}, \textcolor{cyan}{40}, \textcolor{cyan}{41}] are utilized to learn image smoothing operators, which indirectly achieve image detail enhancement by fitting a smoothing layer of the image. There is no doubt that such methods have several inherent drawbacks. First, this task lacks ground truth supervised signals, and some algorithms, namely [\textcolor{cyan}{39}, \textcolor{cyan}{41}], use the outputs of existing algorithms as ground truth supervised signals, which limits the performance of the algorithms to some extent. Second, for each different filter, the deep neural network needs to be retrained accordingly or the pre-trained parameters need to be fine-tuned manually and carefully to achieve the best results, which is undoubtedly time-consuming and labor-intensive.

Nevertheless, deep-learning algorithms have made great strides. In this experiment, the compared algorithms are those that have been popular in the last few years. First, deep neural networks are found to have a low impedance to natural signals and a high impedance to noisy signals naturally, and this property is summarized as the Deep Image Prior (DIP) [\textcolor{cyan}{38}]. With this prior information, a pre-trained neural network can achieve the detail enhancement task of a single image. Second, some non-adaptive image processing tasks, namely the tasks with parameters to be set, [\textcolor{cyan}{39}] dynamically set these parameters utilizing Decoupled Learning (DL). DL borrows the idea of meta-learning to automatically adjust the weights of the pre-trained network through a weighting network, thus enabling the model to be adapted to various applications, including image detail enhancement.  Third, the Contrast Semantic Guided Image Smoothing Network (CSGIS-Net) [\textcolor{cyan}{40}] is designed to facilitate image smoothing by combining a contrast prior and a semantic prior. The supervised signal is enhanced by using undesired smoothing effects as negative teachers and by incorporating segmentation tasks to encourage semantic uniqueness.

\begin{figure}[htbp]	
	\centering
	\includegraphics[scale=0.75]{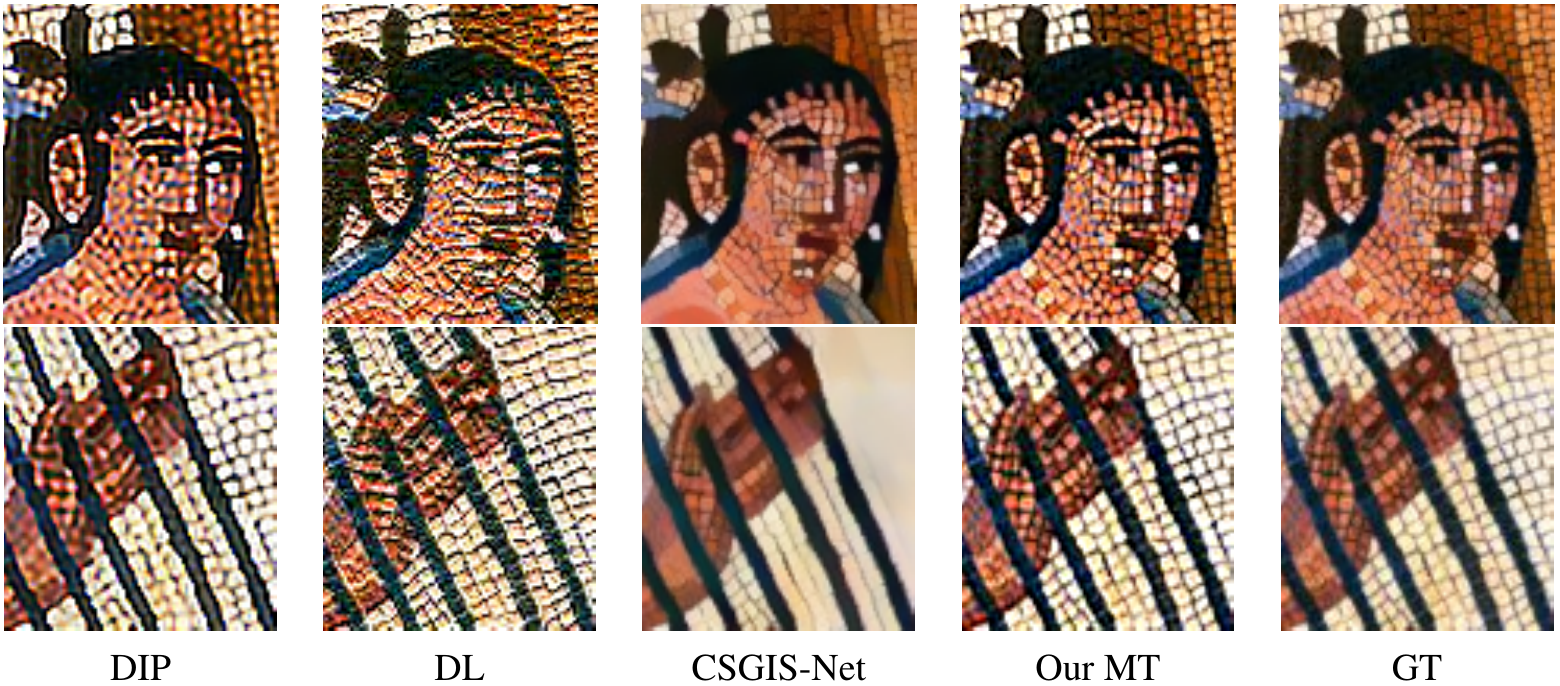}
	\caption{The first visual comparison of the deep learning-based methods. Each method's name is marked below the corresponding image. GT means the Ground Truth (input image) and detail layer magnification factor $ \alpha = 4 $.}
\end{figure}

\begin{figure}[htbp]	
	\centering
	\includegraphics[scale=0.76]{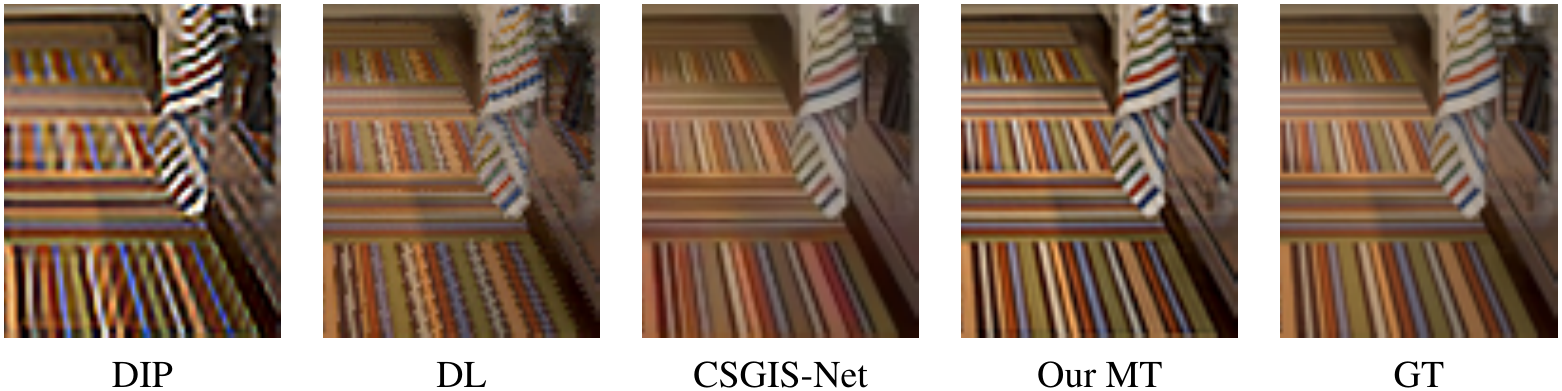}
	\caption{The second visual comparison of the deep learning-based methods. Each method's name is marked below the corresponding image. GT means the Ground Truth (input image) and detail layer magnification factor $ \alpha = 4 $.}
\end{figure}

\begin{figure}[h]	
	\centering
	\includegraphics[scale=0.76]{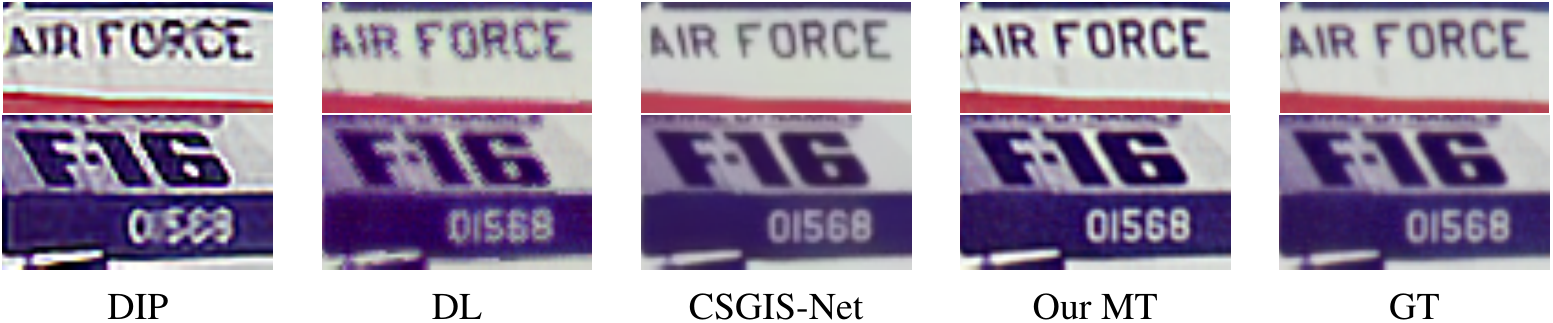}
	\caption{The third visual comparison of the deep learning-based methods. Each method's name is marked below the corresponding image. GT means the Ground Truth (input image) and detail layer magnification factor $ \alpha = 4 $.}
\end{figure}

\begin{figure}[h]	
	\centering
	\includegraphics[scale=0.76]{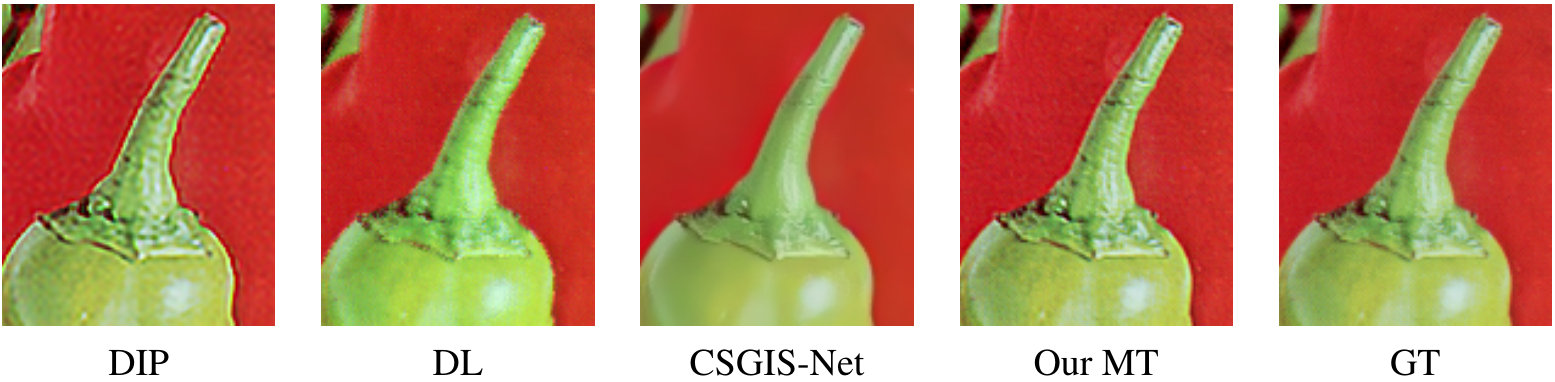}
	\caption{The fourth visual comparison of the deep learning-based methods. Each method's name is marked below the corresponding image. GT means the Ground Truth (input image) and detail layer magnification factor $ \alpha = 4 $.}
\end{figure}

\textcolor{cyan}{Fig. 11} $ \sim $ \textcolor{cyan}{Fig. 14} show four sets of detail-enhanced images based on deep learning algorithms. From these sets of images, we can see that there are many obvious flaws in the results of DIP [\textcolor{cyan}{38}], DL [\textcolor{cyan}{39}], and CSGIS-Net [\textcolor{cyan}{40}], such as distortion artifact, gradient reversal artifact, staircase noise, and blur artifact. First of all, distortion artifact is distinct in the results of DIP [\textcolor{cyan}{38}], such as the carpet texture in  \textcolor{cyan}{Fig. 12} and the figures on the airplane in  \textcolor{cyan}{Fig. 13} are distorted to different degrees. Gradient reversal artifact is also present in the results of DIP [\textcolor{cyan}{38}], i.e. the pixels around the pepper stick in  \textcolor{cyan}{Fig. 14} is very obvious. In addition, staircase noise often appears in the DL [\textcolor{cyan}{39}] results, such as the carpet and towel in  \textcolor{cyan}{Fig. 12}, and the numbers on the cabin in  \textcolor{cyan}{Fig. 13}, where there is lots of staircase noise surrounding their edges. Finally, a blur artifact exists in the CSGIS-Net [\textcolor{cyan}{40}] results. For instance, the mosaic next to the strings in  \textcolor{cyan}{Fig. 11} and the pepper sticks in  \textcolor{cyan}{Fig. 14} that are incorrectly smoothed and blurred. In contrast, our algorithm can generate enhanced images with clear and accurate textures and realistic details, and our proposed algorithm has superiority in visual performance.

To further demonstrate the performance of MT, the above mentioned deep network-based algorithms are tested quantitatively, and the results are shown in \textcolor{cyan}{Fig. 15} and \textcolor{cyan}{Fig. 16}. In \textcolor{cyan}{Fig. 15}, MT achieves the second place in the RMSE metrics test for the natural dataset and the first place in the medical dataset. It is also worth mentioning that MT comes out on top in all SSIM metrics tests, which once again proves that MT effectively protects the structural information of the images from distortion while detail enhancement. In the histogram of \textcolor{cyan}{Fig. 16}, MT basically achieved the top two results, which corroborates the outstanding visual performance ability of MT, and also indirectly shows that MT's generalization ability is very strong, i.e. it is robust to different kinds of textures. Besides, it should be emphasized that the intensity curve in \textcolor{cyan}{Fig. 16} also supports the extraordinary edge preservation ability of the MT algorithm.

\begin{figure}[h]	
	\centering
	\includegraphics[scale=0.35]{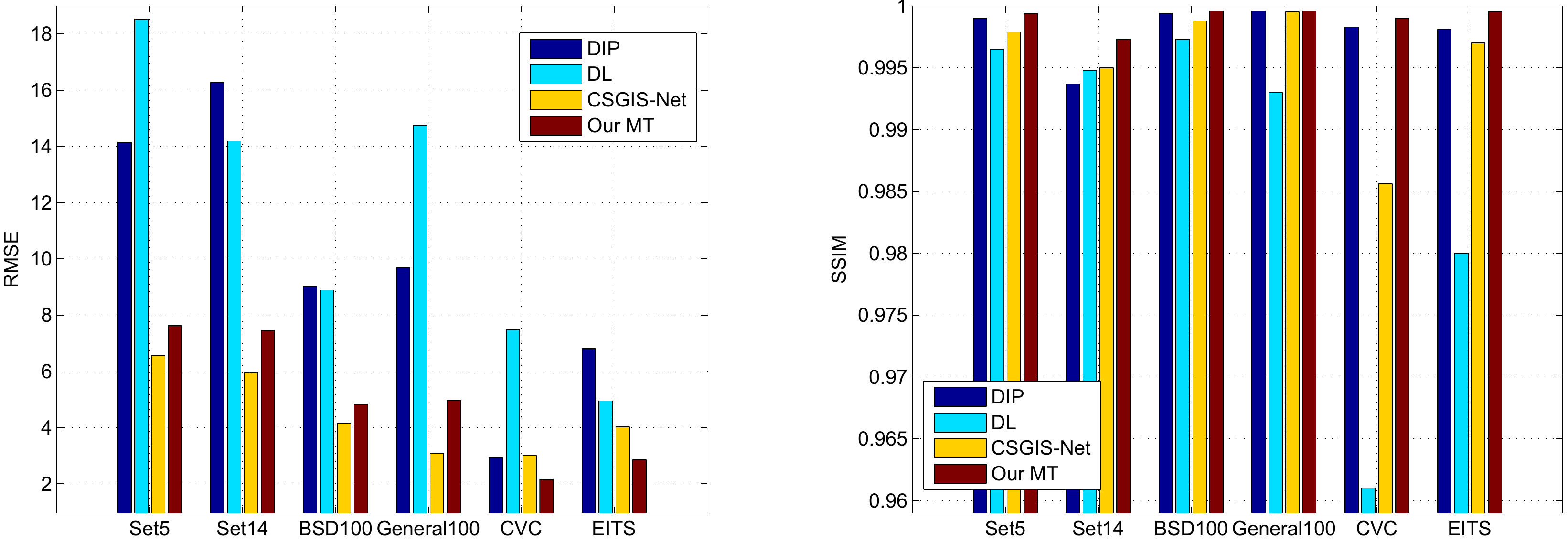}
	\caption{Quantitative testing histograms of deep learning-based algorithms with RMSE and SSIM as test metrics, and detail layer magnification factor $ \alpha = 4 $.}
\end{figure}

\begin{figure}[h]	
	\centering
	\includegraphics[scale=0.35]{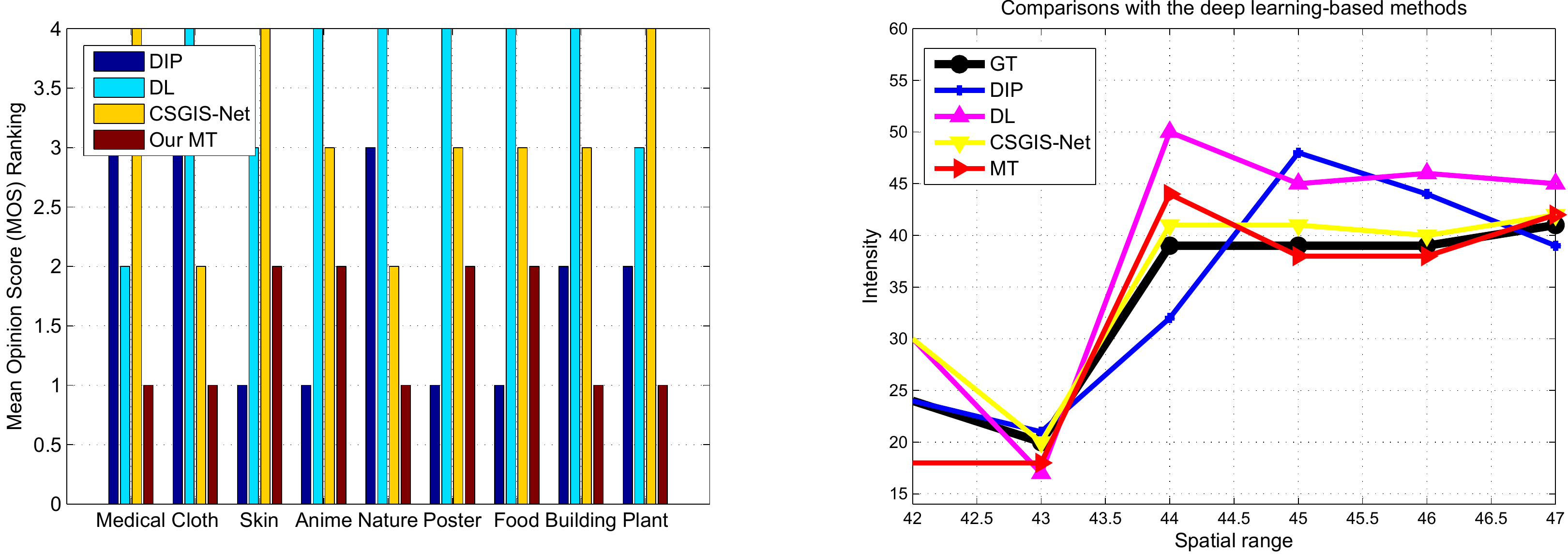}
	\caption{MOS metrics scoring ranking histogram and visualization of intensity curves of the deep learning-based algorithms, and detail layer magnification factor $ \alpha = 4 $.}
\end{figure}

\subsection{Circuit implementation complexity analysis}

\textcolor{cyan}{Table 4} shows the difficulty of implementing different detail enhancement algorithms on the circuit. In general, hardware is a large-scale integrated circuit, which is composed of modules based on high and low-voltage units. For high and low voltages, the simplest operations are addition, subtraction, and integer multiplication. It is not suitable for floating point arithmetic, optimization equation solving, neural network updating, and so on. Local filters contain many floating-point divisions, global filters contain numbers of unsolved optimization equations, and deep learning-based algorithms contain weights update of neural networks, thus these conditions themselves limit the simplicity and possibility of the circuit implementation. However, the detail enhancement algorithms based on residual learning only contain addition, subtraction, searching and patch matching, which is theoretically easy to implement in circuits, that is, our algorithm has lots of practical value. It is worth mentioning that the patch matching operation of our algorithm is relatively independent, which is suitable for using GPU to accelerate the algorithm, and this will be of interest to us in the future.

\begin{table}[htbp]
	\centering
	\caption{Comparisons of circuit implementation complexity among these SOTA algorithms.}
	\begin{small}
		\setlength{\tabcolsep}{1.2mm}{
			\begin{tabular}{c|c|c} \hline
				Categories   &  Method names  & Circuit implementation complexity  \\\hline
				\multirow{3}{*}{Local filter} & GIF [\textcolor{cyan}{2}]  GGIF [\textcolor{cyan}{3}]  &  \multirow{3}{*}{Medium} \\
				\multirow{3}{*}{}  & RGIF[\textcolor{cyan}{7}], EGIF[\textcolor{cyan}{8}]  &  \multirow{3}{*}{} \\
				\multirow{3}{*}{}  & SPGIF[\textcolor{cyan}{15}]  &  \multirow{3}{*}{} \\\hline
				\multirow{3}{*}{Global filter}  & WLS [\textcolor{cyan}{11}], FS [\textcolor{cyan}{13}]  &  \multirow{3}{*}{Hard} \\
				\multirow{3}{*}{}  & BFLS [\textcolor{cyan}{21}], ILS [\textcolor{cyan}{10}] &  \multirow{3}{*}{} \\
				\multirow{3}{*}{}  & TH [\textcolor{cyan}{33}]  &  \multirow{3}{*}{} \\\hline
				\multirow{2}{*}{Residual learning} & LSE [\textcolor{cyan}{27}], ZF [\textcolor{cyan}{9}] &  \multirow{2}{*}{Easy} \\
				\multirow{2}{*}{} & IPRH [\textcolor{cyan}{6}], Our MT &  \multirow{2}{*}{} \\\hline
				\multirow{2}{*}{Deep learning} & VDCNN [\textcolor{cyan}{41}], DIP [\textcolor{cyan}{38}]  & \multirow{2}{*}{Hard}\\
				\multirow{2}{*}{} &  DL [\textcolor{cyan}{39}], CSGIS-Net [\textcolor{cyan}{40}]& \multirow{3}{*}{}\\\hline
		\end{tabular}}
	\end{small}	
\end{table}

\subsection{Ablation study}
The optimization of our system is performed in a way that references the cooling process of the thermodynamic system, so the parameters in the model related to the thermodynamic system are set globally based on the Metropolis theorem. In other words, the ablation study of the parameters unfolds only on the remaining parameters, i.e.  $ \eta $ and $ \mu $, which are all related to the energy function $ E(\mathbf{x}) $. To make the gradient and texture feature more beneficial, ablation studies are conducted on the parameters using a control variable approach.

\begin{figure}[h]	
	\centering
	\includegraphics[scale=0.5]{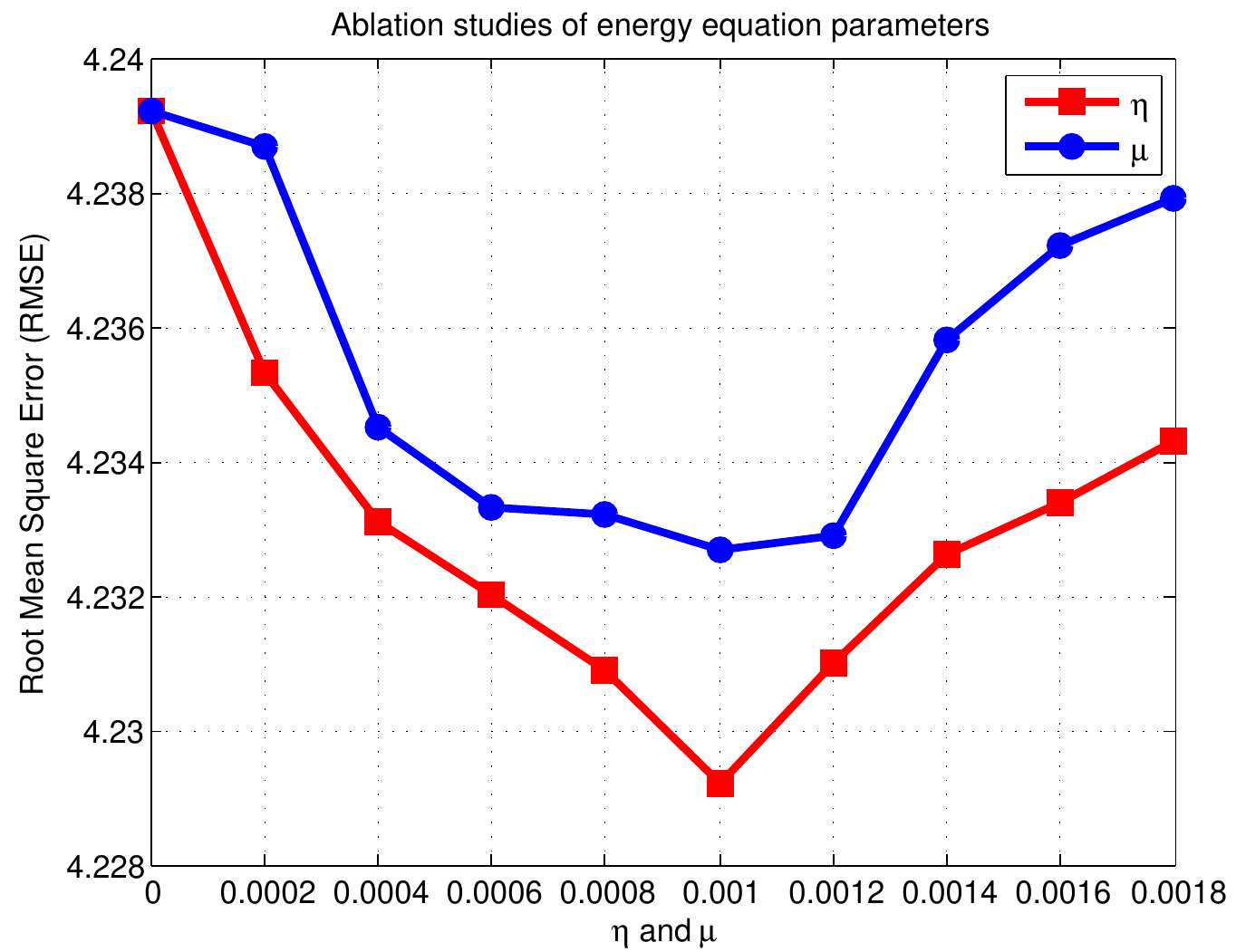}
	\caption{Ablation studies of parameters $ \eta $ and $ \mu $ of the energy function $ E(\mathbf{x}) $.}
\end{figure}

The test image used for the ablation study is img\_069.png, which is taken from the dataset named General100[\textcolor{cyan}{34}], and the evaluation metric used is the Root Mean Square Error (RMSE) mentioned above. In the experiments, the values of parameters $ \eta $ and $ \mu $ are found to remain essentially the same when the system converges to the optimal state, even if other images are used. The curves in \textcolor{cyan}{Fig. 17} show the final results. From \textcolor{cyan}{Fig. 17}, it can be concluded that the RMSE value of the system is relatively high when $ \eta $ and $ \mu $ are set to 0. This indicates that both gradient and texture features have their contributions to the model of image detail enhancement, and one cannot be separated from the other. In addition, when the values of $ \eta $ and $ \mu $ are set to 0.001, the system has the lowest RMSE metric value, and the system converges to the best state at this time, so in our experiments, the values of variables $ \eta $ and $ \mu $ are set to 0.001.

\subsection{Model complexity analysis}
Compared with the detail enhancement algorithm based on patch matching, the detail enhancement algorithm based on the Metropolis theorem has its advantage, which lies in the improvement of searching accuracy. But there are pros and cons to everything, which inevitably leads to an increase in the convergence times of searching. To achieve global convergence, the seeds are initialized and spread randomly so that the searching times are not the same each time.  

For an image with resolution $ s\times t $, it is divided into small patches with resolution $ r\times r $. Suppose that the searching times used by the current small patch to find its best matching small patch is $ N_i $, and the extra searching times due to seed diffusion is $ N_i' $. For the current small patch, the total number of searching times is $ N_i+N_i' $. Suppose that a searching and matching takes $ T_s $, and the previous operation takes $ T_{pre} $,  so the time complexity of the
system is $ \mathcal{O}\left(T_{pre} + T_s\sum\nolimits_{i=1}^{\frac{s\times t}{r\times r}} (N_i+N_i')\right) = \mathcal{O}\left(T_{pre} + T_s\frac{s\times t}{r\times r}\bar{N}\right) \approx \mathcal{O}\left(\xi n^2\right)$, where $ \bar{N} $ is the average number of searching and $ \xi $ is a positive constant.

\subsection{Limitations and future work}
The algorithm proposed in this paper achieves high scores in both visual performance and quantitative tests, but these scores are obtained at the expense of time complexity, and the time complexity of our algorithm is about $ \mathcal{O}\left( n^2\right) $. It runs slowly compared to traditional local filter-based algorithms, and there is no doubt that the proposed algorithm has a lot of room for improvement in terms of running efficiency. In the future, we will modify this algorithm in two ways. First, we will develop a deep-learning version of the algorithm and accelerate it using lightweight deep network techniques. In addition, we will develop a hardware-level version of the algorithm and accelerate this algorithm with GPUs, so that it can make itself more commercially viable by getting a significant speedup while maintaining performance.

\section{Conclusion}
In this paper, a detail enhancement algorithm has been proposed based on the Metropolis theorem. First, a new energy function is minimized to initialize the residual feature. Second, the Metropolis theorem is applied to refine the searching and matching process and find out the features that are more suitable for the detail layer. Finally, the detail-enhanced image is achieved by amplifying the detail layer. It has to be said that our algorithm has an excellent performance in both subjective tests and quantitative data and curves. What’s more, because of the simplicity of the algorithm itself, it is easy to be implemented by the circuit, which has a strong practical value.

\section{Declaration of Competing Interest}
The authors declare that they are not aware of the possibility of competing for financial interests or personal relationships affecting the work reported in this paper.

\section{Fund} This work was supported in part by the National Natural Science Foundation of China under Grant (No.52204177) and supported in part by the Fundamental Research Funds for the Central Universities (2020QN49)

\end{document}